\documentclass{article}

\PassOptionsToPackage{numbers,compress}{natbib}

\usepackage{booktabs}
\usepackage{multirow}
\usepackage{xcolor}
\usepackage{colortbl}
\usepackage{makecell}
\usepackage{array}
\usepackage{mathrsfs}
\usepackage{graphicx}
\usepackage{caption}
\usepackage{amsmath}
\usepackage{amsmath}
\definecolor{headerblue}{RGB}{220, 232, 245}
\definecolor{rowgray}{RGB}{245, 246, 248}
\definecolor{bestgold}{RGB}{255, 243, 205}
\definecolor{midrulecolor}{RGB}{180, 195, 210}

\usepackage{booktabs}
\usepackage{multirow}
\usepackage{xcolor}
\usepackage{colortbl}
\usepackage{makecell}
\usepackage{array}
\usepackage{mathrsfs}
\usepackage{graphicx}

\usepackage{tikz}

\newcommand{\blackcircled}[1]{%
  \tikz[baseline=(char.base)]{
    \node[
      shape=circle,
      fill=black,
      text=white,
      inner sep=0pt,
      minimum size=0.95em,
      font=\scriptsize
    ] (char) {#1};
  }%
}

\definecolor{modernGreen}{HTML}{8BC34A} 
\definecolor{modernOrange}{HTML}{FF9800} 

\newcommand{\greencircled}[1]{%
  \tikz[baseline=(char.base)]{
    \node[
      shape=circle,
      fill=modernGreen,
      text=white,
      inner sep=0pt,
      minimum size=0.95em,
      font=\scriptsize
    ] (char) {#1};
  }%
}

\newcommand{\orangecircled}[1]{%
  \tikz[baseline=(char.base)]{
    \node[
      shape=circle,
      fill=modernOrange,
      text=white,
      inner sep=0pt,
      minimum size=0.95em,
      font=\scriptsize
    ] (char) {#1};
  }%
}

\definecolor{headerblue}{RGB}{220, 232, 245}
\definecolor{rowgray}{RGB}{245, 246, 248}
\definecolor{bestgold}{RGB}{255, 243, 205}
\definecolor{midrulecolor}{RGB}{180, 195, 210}

 \usepackage[preprint]{neurips_2026}

\usepackage[utf8]{inputenc} 
\usepackage[T1]{fontenc}    
\usepackage{hyperref}       
\usepackage{url}            
\usepackage{booktabs}       
\usepackage{amsfonts}       
\usepackage{nicefrac}       
\usepackage{microtype}      
\usepackage{xcolor}         
\usepackage{enumitem}

\title{Scene-SAM3D: Multi-View Scene Asset Generation Without Fine-Tuning}

\author{
  Yuqi Zhang$^{1,3}$ \quad
  Yadan Luo$^{1,\dagger}$ \quad
  Xiangyu Sun$^{1}$ \quad
  Fengyi Zhang$^{1}$ \quad
  Zi Huang$^{1}$ \quad
  Xin Tan$^{2,3,\dagger}$ \\[1mm]
  {\small
  $^{1}$The University of Queensland \quad
  $^{2}$East China Normal University \quad
  $^{3}$Shanghai AI Laboratory}
}

\begin{document}
\maketitle

\makeatletter
\begingroup
\renewcommand{\thefootnote}{}
\renewcommand{\@makefntext}[1]{\noindent #1}
\footnotetext{$^{\dagger}$ Corresponding authors.}
\endgroup
\makeatother



\begin{abstract}
High-quality 3D scene assets are critical for embodied applications such as robotic manipulation, navigation, and simulation. Despite their strong object priors, recent single-image 3D generation models such as SAM3D remain insufficient for real-world scenes, where severe occlusions, redundant observations, and cross-view inconsistencies make reliable scene generation challenging. We introduce \textbf{Scene-SAM3D}, a training-free framework that extends SAM3D from single-view object generation to calibrated multi-view scene asset generation. Scene-SAM3D selects a compact set of complementary views, reducing observation redundancy while providing additional evidence for regions occluded in individual views. Based on the selected views, it performs step-efficient latent velocity fusion to integrate multi-view evidence and suppress cross-view conflicts in canonical space. Finally, a lightweight rigid-object Gaussian optimization refines the scene layout within 200 iterations while preserving the generated object geometry. Experiments on Replica and ScanNet++ demonstrate consistent improvements at both instance and scene levels, with our method reducing scene-level CD by 43.8\% on Replica and 30.9\% on ScanNet++, while cutting flow-model sampling FLOPs and wall-time latency by nearly 20\% under the same multi-view setting. Code will be released at \url{https://github.com/xibi777/Scene-SAM3D}.

\end{abstract}

\section{Introduction}
Faithful and spatially coherent 3D scene assets are essential for realistic digital environments, with applications in immersive AR/VR~\cite{jones2014roomalive, orts2016holoportation, newcombe2011kinectfusion, sra2018oasis, xu2023vrnerf, huang2026litereality} and embodied simulation~\cite{wang2024robogen, gu2023maniskill2, khanna2024hssd200, puig2024habitat3, ma2023sqa3d, majumdar2024openeqa}.
A practical way to obtain such assets is to reconstruct or generate 3D meshes from 2D visual observations.  
Recent foundation models~\cite{trellis, zhao2025hunyuan3d, li2025triposg, sam3d} bridge this gap in part by turning a \emph{single} natural image into a structured object-centric prediction of shape, appearance, and camera-relative pose, with SAM3D~\citep{sam3d} as a prominent example.  Yet realistic scenes are rarely observed from a single view.
Calibrated captures instead produce long multi-view sequences with abundant redundancy, pervasive occlusion, and no obvious ``best'' view for interpreting each instance ahead of inference.

\begin{figure}[t]\vspace{-2ex}\vspace{-2ex}
  \centering
    \includegraphics[width=\linewidth]{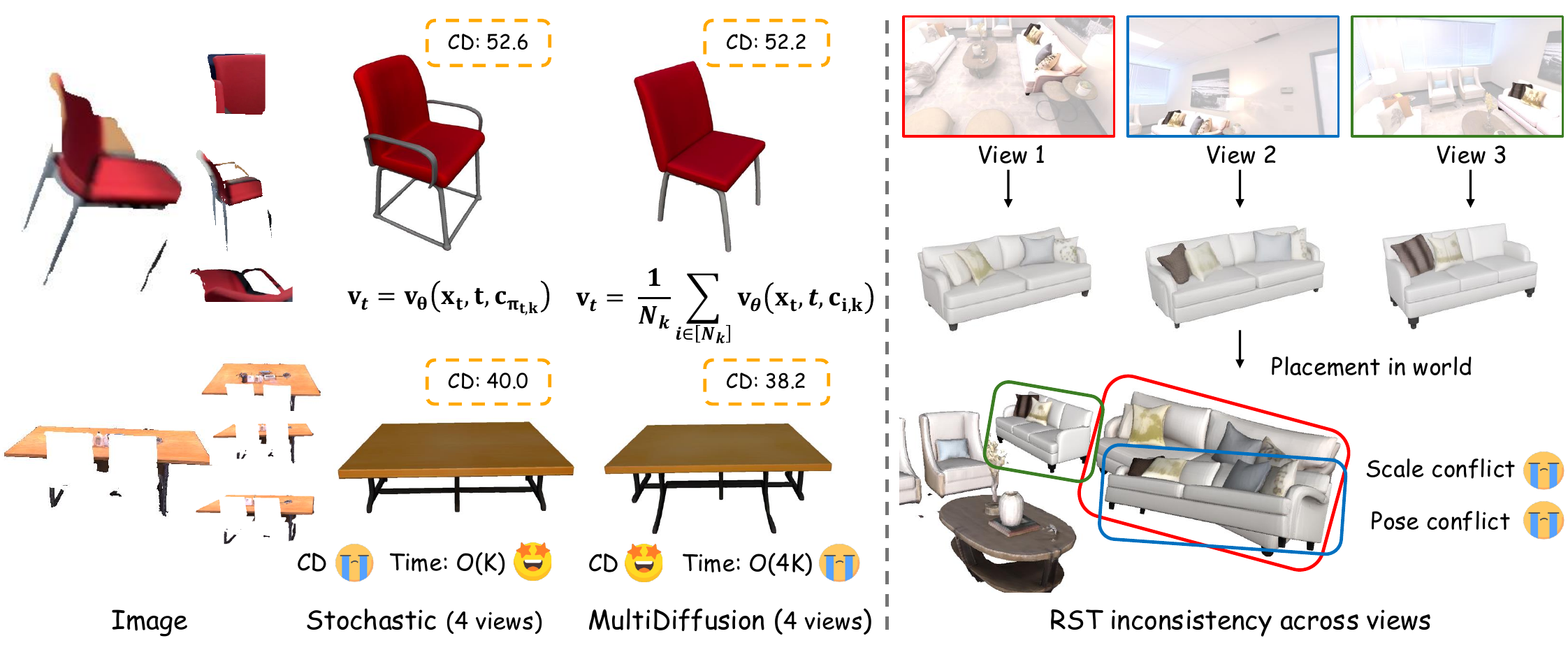}\vspace{-1ex}
  \caption{\textbf{Our preliminary study on extending SAM3D with multi-view fusion.}
Left: stochastic conditioning is efficient but unstable, whereas MultiDiffusion improves geometry at a higher cost. Right: different views produce different predictions of rotation,
scale, and translation.}\label{fig:pre}\vspace{-3ex}
\end{figure}

Generating scene assets under these conditions is therefore a genuinely \textit{multi-view} problem: for each foreground object we seek a complete mesh from complementary multi-view observations, but we simultaneously require that all instances be placed compatibly into a shared world frame consistent with multi-view evidence.
To support multi-view conditioning, two common strategies are stochastic view
conditioning and multi-view velocity fusion. Stochastic view conditioning \cite{trellis}, which samples different views during generation, provides a simple way to use multiple observations but does not explicitly reconcile them across views. Conversely, multi-view velocity method such as MultiDiffusion \cite{multidiffusion}, which averages velocity fields induced by multiple views each step usually improves reconstruction quality over stochastic conditioning \cite{reconviagen}. However, even multiple observations may not fully cover an
object, leaving missing geometry or inducing hallucinated structures in unseen
regions as shown in Fig.~\ref{fig:pre} (left). Moreover, multi-view fusion introduces an additional geometry conflict:
it implicitly assumes that per-view updates are compatible in a shared canonical
space. When this assumption fails, direct averaging can produce duplicated or layered structures, e.g., the duplicated table legs in Fig.~\ref{fig:pre} (left).

Multi-view inference also introduces pose conflicts. For the same instance, different views may produce substantially different predictions of rotation,
scale, and translation. Once transformed into the shared world frame, these
discrepancies lead to incompatible object sizes, orientations, and placements, as
shown in Fig.~\ref{fig:pre} (right). More importantly, the view most favorable for
mesh quality is often not the most reliable for downstream layout estimation.
This mismatch motivates decoupling object-shape reconstruction from scene-layout
refinement.

These observations suggest that multi-view scene generation should be treated as a structured \textbf{selection--fusion--alignment} problem rather than as direct multi-image conditioning.
First, redundant frames should be filtered out while retaining views that provide complementary 3D evidence.
Second, multi-view fusion in a flow-based generator must account for canonical-space disagreement, since uniformly averaging view-conditioned velocity updates can mix incompatible shape hypotheses.
Third, after object-level reconstruction, scene-level layout can benefit from the full trajectory of calibrated views, which provides stronger constraints on object placement than any single reconstruction view alone.
Together, these considerations motivate a pipeline that selects complementary views, fuses them with conflict awareness, and refines object layout through calibrated multi-view feedback.


We instantiate this design in \textbf{Scene-SAM3D}, a training-free pipeline built on frozen SAM3D.
Following the selection--fusion--alignment decomposition, our method first performs \emph{Occupancy-Coverage View Selection}, which selects an anchor view and complementary helper views by measuring newly observed object occupancy in the shared world frame.
The selected views are then integrated by \emph{Step-Efficient Conflict-Aware Multiview Fusion}, which concentrates multi-view fusion in early sampling steps and suppresses incompatible velocity updates through anchor-guided sign consensus.
Finally, \emph{Rigid-Object Gaussian Layout Refinement} fixes the reconstructed object assets and optimizes their similarity transforms with calibrated multi-view rendering feedback, improving scene-level spatial coherence without modifying the generated meshes.

Without finetuning SAM3D, Scene-SAM3D consistently outperforms single-view and MultiDiffusion fusion baselines. At the instance level, it reduces Chamfer Distance (CD) by 5.36\% over single-view SAM3D on Replica and improves Completeness by 6.98\% on ScanNet++; compared with SAM3D using MultiDiffusion fusion, it further reduces CD by 3.93\%/1.97\% on Replica/ScanNet++ while saving nearly 20\%  flow-model FLOPs and wall-time latency under ten-view inference. At the scene level, it reduces CD by 43.75\%/30.91\% over SAM3D with ICP alignment on Replica/ScanNet++.

Our contributions are summarized as follows.
\begin{itemize}[leftmargin=1.5em]
\item We formulate calibrated multi-view scene asset generation as a coupling between instance-centric flow estimation and globally consistent placement, motivating selection, fusion, and alignment stages that interoperate cleanly with pretrained single-view generators.

\item We introduce Scene-SAM3D, a training-free method combining occupancy-based anchor--helper selection for 3D-complementary redundancy reduction, early-stop multi-view latent fusion with anchor-guided sign consensus to mitigate canonical-space conflicts, and rigid-object Gaussian layout refinement for calibrated multi-view silhouette alignment without editing generated meshes.

\item We evaluate on Replica and ScanNet++, demonstrating consistent gains in object-level metrics and stronger scene-scale reconstruction when refining poses with Gaussian objectives, alongside reduced flow-model inference steps relative to exhaustive multi-view velocity fusion.
\end{itemize}

\section{Preliminaries and Problem Setup}\label{sec:preliminary}
\noindent\textbf{SAM3D.} Before introducing our multi-view extension, we briefly review the base model SAM3D. SAM3D adopts a two-stage flow-matching architecture, where each stage consists of a conditional flow model followed by a task-specific decoder. For each object $k$, given an input RGB image $\mathbf{I}$ and an instance mask $\mathbf{M}$, the Stage 1 Geometry Model defines a conditional generative process $p(\mathbf{O}_k, \mathbf{T}^{w \leftarrow o}_k \mid \mathbf{I}, \mathbf{M})$ for coarse shape and object placement. At inference time, it generates a canonical occupancy grid $\mathbf{O}_k \in \mathbb{R}^{64 \times 64 \times 64}$ and predicts an object-to-world transform $\mathbf{T}^{w \leftarrow o}_k=(\mathbf{R}_k,\mathbf{s}_k,\mathbf{t}_k)$, where $\mathbf{R}_k$, $\mathbf{s}_k$, and $\mathbf{t}_k$ denote the rotation, scale, and translation of object $k$, respectively. Given a calibrated camera pose $C_i$, the predicted camera-relative transform can be lifted to the world frame. Conditioned on $\mathbf{O}_k$, the Stage 2 Texture and Refinement Model defines $p(\mathbf{Z}_k \mid \mathbf{I}, \mathbf{M}, \mathbf{O}_k)$ and generates a refined asset representation $\mathbf{Z}_k$ that encodes refined geometry and texture. Task-specific decoders then convert $\mathbf{Z}_k$ into the final textured asset $\mathcal{G}_k$, such as a Gaussian or mesh representation.

\noindent\textbf{Problem Definition.} We consider \textit{scene-level} asset generation from a calibrated video sequence containing $K$ foreground objects and $N$ views. 
For each object $k$, we define its object-centric observation set as
$\mathcal{D}_k=\{(\mathbf{I}_{i},\mathbf{M}_{i},\mathbf{C}_{i},\mathbf{K})\}_{i=1}^{N_k}$,
where $N_k$ denotes the number of valid observations, and $\mathbf{I}{i}$, $\mathbf{M}{i}$, $\mathbf{C}{i}$, and $\mathbf{K}$ denote the RGB view, instance mask, camera-to-world pose, and shared camera intrinsics, respectively.
Given these observations, our \textbf{goal} is to generate a complete scene asset $\mathcal{S}=\{(\mathcal{G}_k, \mathbf{T}^{w \leftarrow o}_k)\}_{k=1}^K$ and estimate its placement in the world coordinate system. 
We denote the object-to-world transform as $\mathbf{T}^{w \leftarrow o}_k=(\mathbf{R}_k,\mathbf{s}_k,\mathbf{t}_k)$, where $\mathbf{R}_k \in \operatorname{SO(3)}$, $\mathbf{s}_k \in \mathbb{R}^{3}_{+}$, and $\mathbf{t}_k \in \mathbb{R}^{3}$. 
Compared with isolated instance generation, scene-level generation is substantially more demanding: objects are observed through \textit{partial} and often \textit{occluded} views, requiring multi-view evidence for completion, while all recovered assets must be \textit{placed consistently} in a shared world frame. 

\noindent\textbf{Multiview Instance Mesh Generation.} Modern image-to-3D models~\cite{trellis,sam3d} reconstruct plausible geometry and texture from a single image, but effectively leveraging multiple observations remains underexplored. A simple extension is \blackcircled{1} \textit{Stochastic view conditioning}~\cite{trellis}. 
For a flow-based generator with conditional velocity $\mathbf{v}_{\theta}$ and intermediate 3D latent state $\mathbf{x}_t$, one may sample a valid view $\pi_{t,k} \sim q(\cdot \mid \mathcal{D}_k)$ at each generation step and update the latent state by $\mathbf{x}_{t+\Delta t}=\mathbf{x}_{t}+\Delta t \cdot \mathbf{v}_{\theta}(\mathbf{x}_t,t,\mathbf{c}_{\pi_{t,k}})$, where $\mathbf{c}_{\pi_{t,k}}$ denotes the corresponding condition. This strategy is computationally cheap because each step uses only \textit{one} view condition. 
Yet it does not explicitly aggregate cross-view evidence, so the generation trajectory may be driven by incompatible single-view interpretations across timesteps. A more direct alternative is \blackcircled{2} \textit{MultiDiffusion velocity fusion}~\cite{multidiffusion}, where all selected views contribute to the update through $\mathbf{x}_{t+\Delta t}=\mathbf{x}_{t}+\Delta t \cdot \frac{1}{N_k}\sum_{i \in [N_k]}\mathbf{v}_{\theta}(\mathbf{x}_t,t,\mathbf{c}_{i,k})$. 
This formulation improves cross-view geometric consistency and object completeness, but introduces two limitations in scene settings. First, the generation stage is dominated by repeated evaluations of the flow model. Therefore, applying MultiDiffusion fusion makes the inference cost scale with both the number of objects $K$ and the number of selected views $N_k$, approximately as $T\sum_{k=1}^{K}N_k\operatorname{Cost}(\mathbf{v}_{\theta})$, where $T$ denotes the total number of flow inference steps across all generation stages.
Second, uniform fusion assumes that all view conditions are compatible. 
When different views induce different canonical object states, averaging their velocities can merge inconsistent modes and produce duplicated or layered geometry. 

\begin{figure}[t]\vspace{-2ex}\vspace{-2ex}
  \centering
    \includegraphics[width=\linewidth]{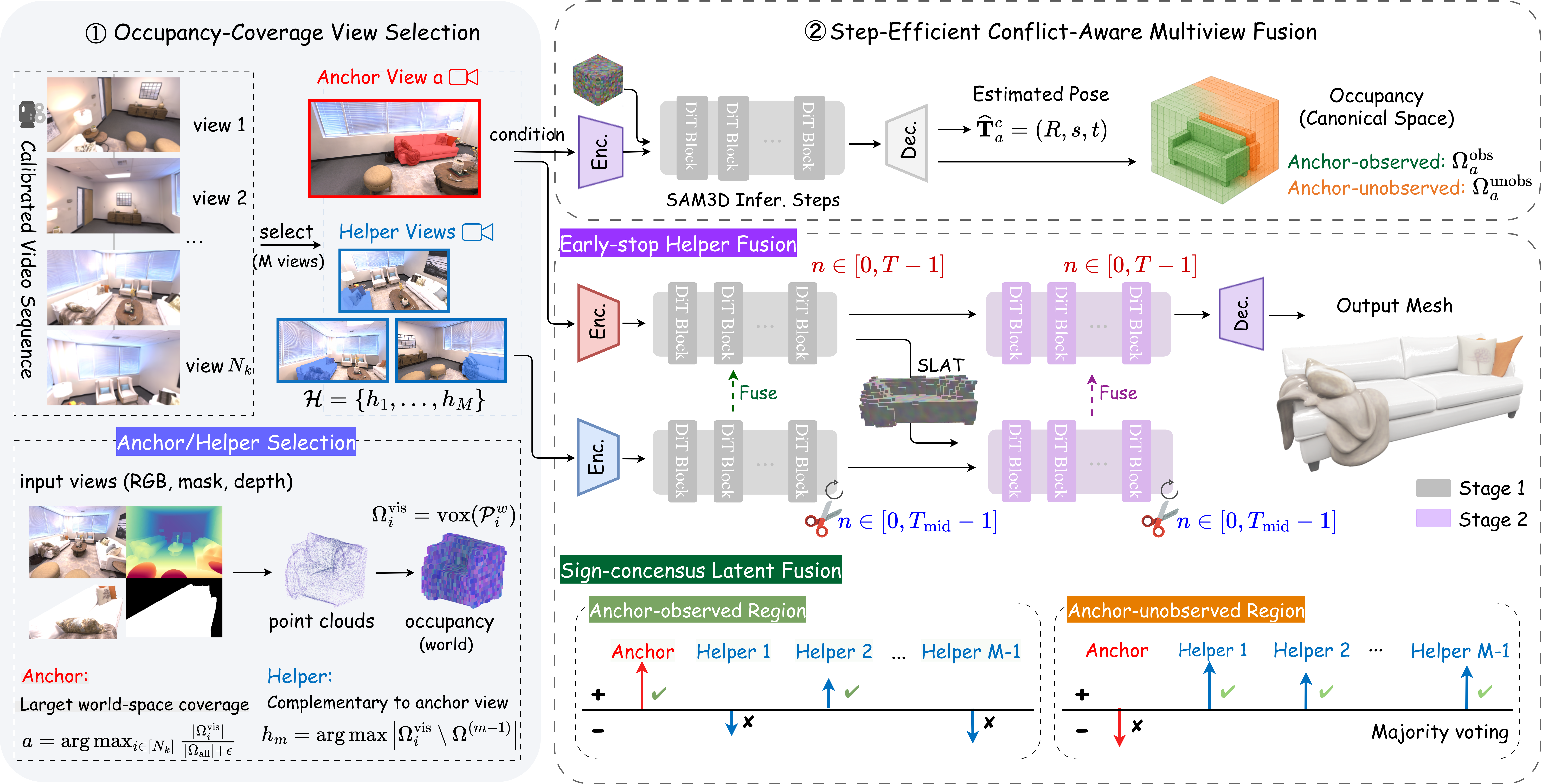}
  \caption{
\textbf{Overview of Scene-SAM3D view selection and multi-view fusion.}
Given calibrated multi-view observations, Scene-SAM3D first selects a compact anchor-helper view set using world-space occupancy coverage. 
Based on the selected views, its performs step-efficient conflict-aware multiview fusion: the anchor view provides the main generation trajectory, while helper views are early-stopped and fused to inject complementary multi-view cues. 
Sign consensus further filters inconsistent latent updates in anchor-observed regions and aggregates reliable evidence in anchor-unobserved regions.
}\vspace{-2ex}
  \label{fig:overview}
\end{figure}

\section{Proposed Approach}\vspace{-1ex}\vspace{-1ex}
As illustrated in Fig.~\ref{fig:overview} and Fig.~\ref{fig:gs}, Scene-SAM3D follows a selection--fusion--alignment pipeline for multi-view scene asset generation. Given calibrated observations, it first selects an anchor-helper view set ($\S$\ref{sec:ocvs}), where helper views complement anchor-unobserved regions through world-space occupancy coverage. It then performs step-efficient conflict-aware latent fusion ($\S$\ref{sec:scmf}) to integrate early multi-view cues while suppressing inconsistent updates. Finally, rigid-object Gaussian layout optimization ($\S$\ref{sec:gaussian_layout}) improves scene coherence without altering object geometry or appearance. We describe the first two components for a fixed object $k$ and omit the index when clear.
\vspace{-1ex}\vspace{-1ex}
\subsection{Occupancy-Coverage View Selection}
\label{sec:ocvs}
Using all $N_k$ observations is both redundant and expensive. 
We therefore select a compact view set consisting of one anchor view $a$ and $M$ helper views $\mathcal{H}=\{h_1,\ldots,h_M\}$, denoted as $\mathcal{Q}=\{a\}\cup\mathcal{H}$. 
The anchor provides the primary appearance and canonical reference, while helper views are selected only to complement object regions that are weakly observed or invisible from the anchor.

\noindent\textbf{Anchor View Selection.} To measure view coverage in 3D rather than in the image plane, we project each masked observation into the world coordinate system. 
Given an estimated depth map $\mathbf{D}_i$, the world-space point set of view $i$ is
\begin{equation}
    \mathcal{P}^{w}_i
    =
    \left\{
    \mathbf{C}_i
    \begin{bmatrix}
    \mathbf{D}_i(\mathbf{u})\mathbf{K}^{-1}\bar{\mathbf{u}} \\
    1
    \end{bmatrix}
    \;\middle|\;
    \mathbf{M}_i(\mathbf{u})=1
    \right\},
\end{equation}
where $\bar{\mathbf{u}}$ is the homogeneous image coordinate and $\mathbf{C}_i$ is the camera-to-world pose. 
We voxelize $\mathcal{P}^{w}_i$ in the shared world frame and obtain the visible occupancy of view $i$ as $\Omega_i^{\mathrm{vis}}=\operatorname{vox}(\mathcal{P}^{w}_i)$. 
The per-view 3D coverage score is defined as the \textit{volume }of the merged occupancy and thus anchor selection reduces to choosing the view with the largest world-space occupied volume:
\begin{equation}\label{eq:O_vis}
\begin{aligned}
    c_i &=
    \frac{|\Omega_i^{\mathrm{vis}}|}{|\Omega_{\mathrm{all}}|+\epsilon},\quad \Omega_{\mathrm{all}}=  
    \bigcup_{i=1}^{N_k}\Omega_i^{\mathrm{vis}}, &a = \arg\max_i c_i,
\end{aligned}
\end{equation}
where $|\Omega_{\mathrm{all}}|$ is shared by all views. Notably, this criterion \textit{differs} from selecting the largest 2D mask $|\mathbf{M}_i|$: a close-up view may contain many object pixels, but after voxelization, repeated samples from the same local surface occupy only a small number of voxels. 
Thus, occupancy coverage favors views that explain a broader 3D extent of the object rather than views with merely larger image footprints.

\noindent\textbf{Helper View Selection.} After selecting the anchor, we greedily add helper views that contribute new 3D evidence beyond the regions already covered. 
Starting from $\Omega^{(0)}=\Omega_a^{\mathrm{vis}}$, the $m$-th helper view is selected by maximizing its complementary occupancy:
\begin{equation}
\begin{aligned}
    h_m
    =
    \arg\max_{i\notin\{a,h_1,\ldots,h_{m-1}\}}
    \left|
    \Omega_i^{\mathrm{vis}}
    \setminus
    \Omega^{(m-1)}
    \right|, \quad
    \Omega^{(m)}
    =
    \Omega^{(m-1)}
    \cup
    \Omega_{h_m}^{\mathrm{vis}} .
\end{aligned}
\end{equation}
In this way, the selected view set separates view reliability from view quantity: the anchor stabilizes canonical generation, while helper views are introduced only when they provide genuinely complementary 3D support. Examples of the selected anchor and helper views are provided in Fig.~\ref{fig:sup_selected_views}.

\vspace{-1ex}
\subsection{Step-Efficient Conflict-Aware Multiview Fusion}
\label{sec:scmf}
The selected view set $\mathcal{Q}=\{a\}\cup\mathcal{H}$ reduces view redundancy, but does not resolve how these views should be fused. 
While \textit{multiview velocity fusion} (Sec.~\ref{sec:preliminary}) can aggregate complementary cues, applying it throughout the full trajectory is both \textit{costly} and \textit{unstable}. It costs $\mathcal{O}((M+1)T)$ conditional evaluations per object, and different view conditions may induce misaligned canonical geometries. 
Consequently, direct averaging can mix incompatible velocity updates and produce duplicated or layered structures. To address these issues, we introduce two strategies as follows:

\noindent\textbf{Early-stop Helper Fusion.} Let $\mathbf{x}_n$ denote the intermediate latent state at $n$-th inference step, where $n=0,1, ...,T-1$.  Let $t_n \in [0,1]$ be the corresponding flow time used by the sampler. Given a view condition $\mathbf{c}_i$, the conditional velocity is predicted as $\mathbf{v}_{\theta}(\mathbf{x}_n,t_n,\mathbf{c}_i)$
A direct fusion strategy updates the latent using the average velocity from all selected views:
\begin{equation}
    \bar{\mathbf{v}}^{\mathrm{dir}}_n
    =
    \frac{1}{|\mathcal{Q}|}
    \sum_{i\in\mathcal{Q}}
    \mathbf{v}_{\theta}(\mathbf{x}_n,t_n,\mathbf{c}_i),
    \qquad
    \mathbf{x}_{n+1}
    =
    \mathbf{x}_n
    +
    \Delta t_n\cdot\bar{\mathbf{v}}^{\mathrm{dir}}_n .
\end{equation}
Empirically, we observe that multi-view cues are most beneficial during early sampling (refer to Fig. \ref{fig:heatmap}), when the trajectory determines coarse object structure and recovers missing geometry. 
In contrast, fusion applied from the middle to the end of the trajectory can amplify canonical conflicts, as the generator increasingly refines view-specific details. This motivates a \textit{\textbf{step-efficient}} strategy: helper views assist the structure-forming phase, but the final trajectory should be stabilized by the anchor.

We therefore fuse anchor and helper views only before a midpoint $T_{\mathrm{mid}}$, and then continue with anchor-only inference:
\begin{equation}\label{eq:early_stop_fusion}
    \bar{\mathbf{v}}_n
    =
    \begin{cases}
        \operatorname{Fuse}
        \left(
        \left\{
        \mathbf{v}_{\theta}(\mathbf{x}_n,t_n,\mathbf{c}_i)
        \right\}_{i\in\mathcal{Q}}
        \right), 
        & n\leq T_{\mathrm{mid}}-1, \\[1mm]
        \mathbf{v}_{\theta}(\mathbf{x}_n,t_n,\mathbf{c}_a), 
        & n>T_{\mathrm{mid}}-1,
    \end{cases}
    \qquad
    \mathbf{x}_{n+1}
    =
    \mathbf{x}_n+\Delta t_n\cdot\bar{\mathbf{v}}_n .
\end{equation}
In practice, $\operatorname{Fuse}(\cdot)$ is implemented as a lightweight consensus operation described below. The same early-stop fusion rule is applied independently to both Stage 1 and Stage 2 of Scene-SAM3D, with the latent state and velocity instantiated in the corresponding stage-specific latent space. 
Compared with running all $M+1$ selected views for all $T$ steps, early-stop fusion reduces the per-object conditional evaluations from $\mathcal{O}((M+1)T)$ to $\mathcal{O}(T+MT_{\mathrm{mid}})$, while preserving helper-view information during the stage where it has the strongest influence on geometry.

\noindent\textbf{Sign-Consensus Latent Fusion.} 
Early stopping limits the duration of multi-view interaction, but conflicting helper signals can still enter the fused update before $T_{\mathrm{mid}}$. 
We therefore define $\operatorname{Fuse}(\cdot)$ in Eq.~\eqref{eq:early_stop_fusion} as a sign-consensus operator over the structured velocity latent. 
At step $n$, each selected view $i\in\mathcal{Q}$ predicts a conditional velocity in the structured latent grid:
$
    \mathbf{Z}_{i,n}
    =
    \mathbf{v}_{\theta}(\mathbf{x}_n,t_n,\mathbf{c}_i)
    \in
    \mathbb{R}^{16\times16\times16\times d}.
$
For readability, we omit the step index $n$ below and write $\mathbf{Z}_i$ for $\mathbf{Z}_{i,n}$. We first identify which latent cells are directly observed by the anchor. 
SAM3D Stage-1 predicts an anchor camera-relative transform 
$\widehat{\mathbf{T}}^c_a=(\widehat{\mathbf{R}}^c_a,\widehat{\mathbf{s}}^c_a,\widehat{\mathbf{t}}^c_a)$, 
which maps a canonical object point $\mathbf{p}^{o}$ to the anchor camera frame:
\begin{equation}
    \mathbf{p}^{c}_{a}
    =
    \widehat{\mathbf{R}}^c_a
    \left(
    \widehat{\mathbf{s}}^c_a \odot \mathbf{p}^{o}
    \right)
    +
    \widehat{\mathbf{t}}^c_a .
\end{equation}
Thus, each visible anchor pixel $\mathbf{u}\in\{\mathbf{u}\mid \mathbf{M}_{a}(\mathbf{u})=1\}$ can be mapped to canonical object frame by:
\begin{equation}
    \mathbf{p}^{o}
    =
    \left(
    \widehat{\mathbf{s}}^c_a
    \right)^{-1}
    \odot
    \left(
    \widehat{\mathbf{R}}^c_a
    \right)^{\top}
    \left(
    \mathbf{p}^{c}_{a}
    -
    \widehat{\mathbf{t}}^c_a
    \right),
    \qquad
    \mathbf{p}^{c}_{a}
    =
    \mathbf{D}_{a}(\mathbf{u})\mathbf{K}^{-1}\bar{\mathbf{u}}.
\end{equation}
Voxelizing these canonical points on the structured latent grid gives the anchor-observed region $\Omega_a^{\mathrm{obs}}\subseteq\Omega$. 
The remaining region is $\Omega_a^{\mathrm{unobs}}=\Omega\setminus\Omega_a^{\mathrm{obs}}$. 
All sign operations below are applied element-wise over spatial cells and latent channels.

- \textbf{Case \greencircled{A} Anchor-observed region:} For $u\in\Omega_a^{\mathrm{obs}}$, the anchor is treated as the reference because it provides the most reliable local evidence. 
Only proposals whose signs \textit{agree} with the \textit{anchor} are allowed to contribute:
\begin{equation}
    \mathcal{I}_{\mathrm{obs}}(u)
    =
    \left\{
    i\in\mathcal{Q}
    \mid
    \operatorname{sgn}(\mathbf{Z}_i(u))
    =
    \operatorname{sgn}(\mathbf{Z}_a(u))
    \right\}.
\end{equation}
The fused structured velocity is then
\begin{equation}
    \widetilde{\mathbf{Z}}(u)
    =
    \frac{1}{|\mathcal{I}_{\mathrm{obs}}(u)|}
    \sum_{i\in\mathcal{I}_{\mathrm{obs}}(u)}
    \mathbf{Z}_i(u),
    \qquad
    u\in\Omega_a^{\mathrm{obs}} .
\end{equation}
- \textbf{Case \orangecircled{B} Anchor-unobserved region:} For $u\in\Omega_a^{\mathrm{unobs}}$, the anchor no longer has privileged evidence. 
We instead let all selected views vote for the dominant direction:
\begin{equation}
\mathcal{I}_{\mathrm{unobs}}(u)
    =
    \left\{
    i\in\mathcal{Q}
    \mid
    \operatorname{sgn}(\mathbf{Z}_i(u))
    =
    s(u)
    \right\}, 
    s(u)
    =
    \operatorname{sgn}
    \left(
    \sum_{i\in\mathcal{Q}}
    \operatorname{sgn}(\mathbf{Z}_i(u))
    \right),
\end{equation}
and average only the proposals that agree with this majority sign:
\begin{equation}
    \widetilde{\mathbf{Z}}(u)
    =
    \frac{1}{|\mathcal{I}_{\mathrm{unobs}}(u)|}
    \sum_{i\in\mathcal{I}_{\mathrm{unobs}}(u)}
    \mathbf{Z}_i(u).
\end{equation}
Finally, the sign-consensus operator is defined as $\operatorname{Fuse}
    \left(
    \left\{
    \mathbf{v}_{\theta}(\mathbf{x}_t,t,\mathbf{c}_i)
    \right\}_{i\in\mathcal{Q}}
    \right)
    =
    \widetilde{\mathbf{Z}}_t$.
Thus, anchor-visible regions are protected by the most reliable canonical reference, while anchor-unobserved regions are completed through multi-view consensus. This suppresses local canonical conflicts without discarding complementary geometry from helper views.

\begin{figure}[t]\vspace{-2ex}\vspace{-2ex}
  \centering
    \includegraphics[width=\linewidth]{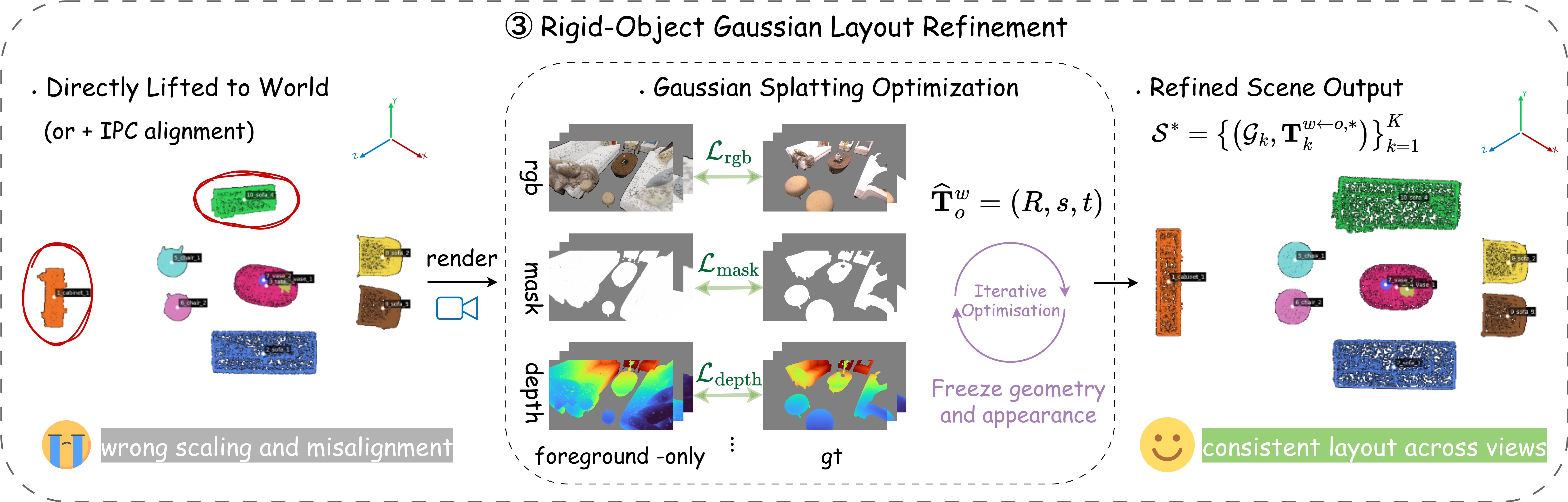}
  \caption{
\textbf{Rigid-object Gaussian layout refinement.}
Scene-SAM3D refines the scene by optimizing object-level similarity transforms in a foreground-only Gaussian representation. 
}\vspace{-2ex}
  \label{fig:gs}
\end{figure}

\subsection{Rigid-Object Gaussian Layout Refinement}\vspace{-1ex}
\label{sec:gaussian_layout}

As shown in Fig.~\ref{fig:gs}, generated objects remain in object-centric coordinates, while the pose and scale lifted from a single view may be inaccurate in the metric world frame.
A simple solution is: (1) direct mask-depth backprojection (BP), or (2) refining the BP with Iterative Closest Point (ICP)~\citep{icp} against masked depth points. However, BP is sensitive to mask and depth noise, and ICP is often under-constrained for truncated or heavily occluded objects.
We therefore introduce a lightweight rigid-object Gaussian optimization that exploits calibrated multiview evidence.
For each eligible generated asset $\mathcal{G}_k$, we construct a downsampled object-centric Gaussian proxy
$\mathcal{P}_k=\{\boldsymbol{\mu}^{o}_{k,j}\}_{j=1}^{J_k}$, where $j$ indexes Gaussian primitives within instance $k$.
The Gaussian geometry, opacity, and appearance of all proxies remain fixed, and the $K$ eligible instances are optimized jointly by updating only their object-level similarity transforms
$\mathbf{T}^{w\leftarrow o}_k=(\mathbf{R}_k,s_k,\mathbf{t}_k)$.
For each instance, the same transform is shared by all its Gaussian primitives; thus, for $j=1,\ldots,J_k$, the transformed world-space mean is
\begin{equation}
\boldsymbol{\mu}^{w}_{k,j}
=
s_k \mathbf{R}_k
\left(
\boldsymbol{\mu}^{o}_{k,j}
-
\mathbf{c}^{o}_{k}
\right)
+
\mathbf{t}_k ,
\end{equation}
where $\mathbf{c}^{o}_{k}$ is the canonical object center.
The Gaussian orientations and scales are transformed accordingly by the same $\mathbf{R}_k$ and $s_k$, while all canonical Gaussian parameters remain unchanged.
For each valid instance-view pair $(k,i)$, given the camera-to-world pose $\mathbf{C}_{k,i}$ and intrinsics $\mathbf{K}$, we render the transformed proxy into view $i$ with a differentiable Gaussian rasterizer, producing an opacity map
$\widehat{\mathbf{A}}_{k,i}(\mathbf{T}^{w\leftarrow o}_k)$.
We also render all optimized instances jointly to obtain scene-level color, depth, and opacity for additional multiview constraints.
The object transforms are optimized with
\begin{equation}
\begin{split}
\mathcal{L}_{\mathrm{layout}}
={}&
\lambda_{m}\mathcal{L}_{\mathrm{mask}}
+
\lambda_{\mathrm{rgb}}\mathcal{L}_{\mathrm{fg}}
+
\lambda_{d}\mathcal{L}_{\mathrm{depth}}
+
\lambda_{b}\mathcal{L}_{\mathrm{bnd}}
\
+
\lambda_{c}\mathcal{L}_{\mathrm{comp}}
+
\lambda_{p}\mathcal{L}_{\mathrm{pen}}
+
\mathcal{L}_{\mathrm{reg}} .
\end{split}
\end{equation}
Here, $\mathcal{L}_{\mathrm{mask}}$, $\mathcal{L}_{\mathrm{fg}}$, $\mathcal{L}_{\mathrm{depth}}$, and $\mathcal{L}_{\mathrm{bnd}}$ supervise instance masks, foreground appearance, depth, and silhouette boundaries, respectively. Meanwhile, $\mathcal{L}_{\mathrm{comp}}$ enforces scene-level rendering consistency, $\mathcal{L}_{\mathrm{pen}}$ penalizes inter-object penetration, and $\mathcal{L}_{\mathrm{reg}}$ regularizes the solution toward the initial layout.
Since the optimization only updates object-level RST parameters rather than Gaussian geometry or appearance, it is lightweight and typically converges within a few hundred iterations. 
The final output is the refined scene asset
\begin{equation}
    \mathcal{S}^{*}
    =
    \left\{
    \left(
    \mathcal{G}_k,
    \mathbf{T}^{w\leftarrow o,*}_k
    \right)
    \right\}_{k=1}^{K}.
\end{equation} 

\begin{figure}[t]\vspace{-2ex}
  \centering
     \includegraphics[width=\linewidth]{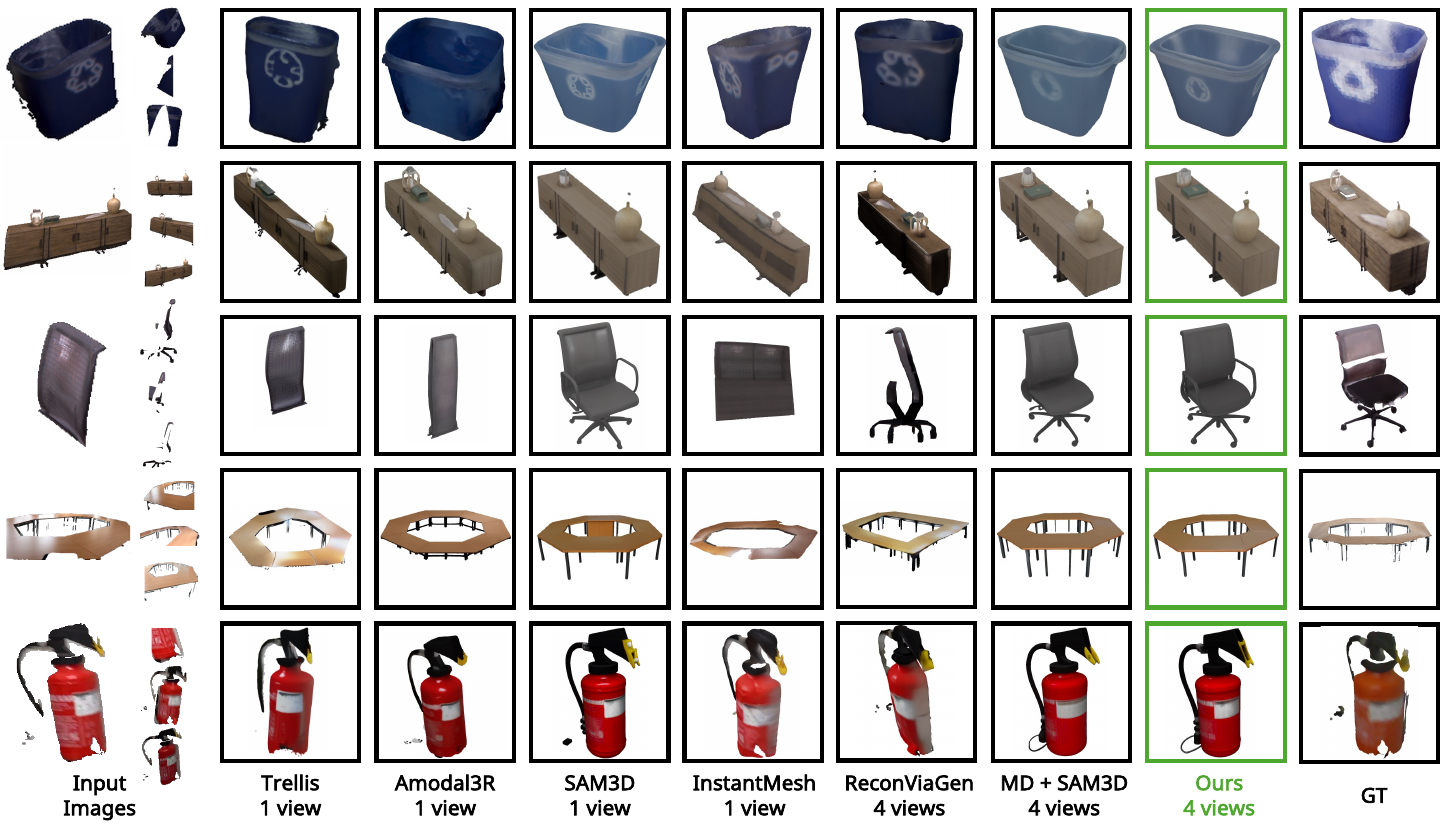}
     \caption{Qualitative instance-level results on the Replica and ScanNet++ datasets.}
    \label{fig:instance}
\end{figure}

\begin{table*}[t]
   \centering
    \caption{
\textbf{Quantitative comparison with single-view methods and reproduced multi-view baselines.}
``random'' denotes random selection, while ``coverage'' denotes our occupancy-coverage strategy.
MD + SAM3D applies MultiDiffusion fusion to SAM3D.
Although InstantMesh takes a single image, it synthesizes multiple intermediate views and is therefore categorized as a multi-view method.}
\vspace{-1ex}
   \label{tab:main_results}
   \setlength{\tabcolsep}{6pt}
   \renewcommand{\arraystretch}{1.25}
   \newcolumntype{C}[1]{>{\centering\arraybackslash}p{#1}}
   \resizebox{\textwidth}{!}{
    \begin{tabular}{ll C{1.0cm} C{1.0cm} C{1.0cm} C{1.3cm} C{1.0cm} C{1.0cm} C{1.0cm} C{1.0cm} C{1.0cm} C{1.0cm} C{1.3cm}}
       \toprule
        \multicolumn{2}{c}{} &
      \multicolumn{7}{c}{\textbf{Replica \cite{straub2019replica}}} &
        \multicolumn{4}{c}{\textbf{ScanNet++ \cite{yeshwanth2023scannetpp}}} \\
       \cmidrule(lr){3-9} \cmidrule(lr){10-13}
       \multicolumn{2}{c}{} &
       \multicolumn{4}{c}{\textit{Geometry}} &
       \multicolumn{3}{c}{\textit{Reconstruction}} &
      \multicolumn{4}{c}{\textit{Geometry}} \\
        \cmidrule(lr){3-6} \cmidrule(lr){7-9} \cmidrule(lr){10-13}
       \textbf{Method} & \textbf{View} &
       CD$^\downarrow$ &
        Comp$^\downarrow$ &
      NC$^\uparrow$ &
       \makecell{F@0.05$^\uparrow$} &
       PSNR$^\uparrow$ &
       SSIM$^\uparrow$ &
       LPIPS$^\downarrow$ &
        CD$^\downarrow$ &
        Comp$^\downarrow$ &
        NC$^\uparrow$ &
        \makecell{F@0.05$^\uparrow$} \\
        \midrule
        \rowcolor{gray!10}
        \multicolumn{13}{l}{\textit{Single-view methods}} \\
        
        \multirow{2}{*}{$\operatorname{Trellis}$ \cite{trellis}}
        & $\operatorname{1 (random)}$   & 114.48 & 93.89 & 58.44 & 0.33 &  9.80 & 0.60 & 0.54 & 143.48 & 128.81 & 54.66 & 0.22 \\
        & $\operatorname{1 (coverage)}$ &  90.41 & 81.83 & 62.99 & 0.42 &  9.97 & 0.63 & 0.57 & 115.41 & 112.13 & 62.00 & 0.32 \\
        
        \addlinespace[0.35em]
        \multirow{2}{*}{$\operatorname{Amodal3R}$ \cite{wu2025amodal3r}}
        & $\operatorname{1 (random)}$   & 73.38 & 70.91 & 66.13 & 0.50 & 10.92 & 0.65 & 0.51 & 102.77 & 85.34 & 61.77 & 0.34 \\
        & $\operatorname{1 (coverage)}$ & 58.87 & 52.46 & 68.70 & 0.58 & 11.11 & 0.64 & 0.48 &  75.78 & 59.19 & 68.58 & 0.48 \\
        
        \addlinespace[0.35em]
        \multirow{2}{*}{$\operatorname{SAM3D}$ \cite{sam3d}}
        & $\operatorname{1 (random)}$   &  61.96 & 55.63 & 68.30 & 0.59 & 12.55 & 0.71 & 0.46 &  79.25 &  60.49 & 67.83 & 0.46 \\
        & $\operatorname{1 (coverage)}$ &  41.60 & 33.20 & \textbf{75.26} & 0.71 & 13.08 & 0.70 & 0.41 &  57.71 &  39.71 & 73.35 & 0.60 \\
        
        \midrule
        \rowcolor{gray!10}
        \multicolumn{13}{l}{\textit{Multi-view methods}} \\
        
        \multirow{2}{*}{$\operatorname{InstantMesh}$ \cite{xu2024instantmesh}}
        & $\operatorname{1 (random)}$   & 93.29 & 105.31 & 62.22 & 0.40 & 11.42 & 0.68 & 0.49 & 111.76 & 112.79 & 58.49 & 0.30 \\
        & $\operatorname{1 (coverage)}$ & 70.92 &  77.31 & 67.27 & 0.50 & 11.78 & 0.69 & 0.53 &  83.44 &  75.77 & 64.77 & 0.41 \\
        
        \addlinespace[0.35em]
        \multirow{2}{*}{$\operatorname{ReconViaGen}$ \cite{reconviagen}}
        & $\operatorname{4 (random)}$   & 81.24 & 89.79 & 64.25 & 0.45 & 9.39 & 0.60 & 0.53 & 109.26 & 114.83 & 57.04 & 0.33 \\
        & $\operatorname{4 (coverage)}$ & 70.76 & 75.15 & 65.62 & 0.53 & 9.42 & 0.60 & 0.51 &  92.80 &  91.56 & 59.59 & 0.41 \\
        
        \addlinespace[0.35em]
        \multirow{2}{*}{$\operatorname{MD}$ \cite{multidiffusion}+$\operatorname{SAM3D}$ \cite{sam3d}}
        & $\operatorname{4 (random)}$   &  45.83 & 37.66 & 74.24 & 0.69 & 13.09 & 0.71 & 0.42 &  60.01 &  38.89 & 73.00 & 0.57 \\
        & $\operatorname{4 (coverage)}$ &  40.98 & 32.06 & 74.86 & 0.72 & 13.19 & 0.71 & 0.40 &  58.51 &  37.57 & 74.07 & 0.60 \\
        
        \addlinespace[0.35em]
        \rowcolor{headerblue}
        $\operatorname{Ours}$ & $\operatorname{4 (coverage)}$
        & \textbf{39.37} & \textbf{30.65} & 75.17 & \textbf{0.73}
        & \textbf{13.20} & \textbf{0.71}
        & \textbf{0.40}
        & \textbf{57.36} & \textbf{36.94} & \textbf{74.50} & \textbf{0.61} \\
        
        \bottomrule

   \end{tabular}\vspace{-2ex}
   }\vspace{-2ex}
  \end{table*}

\vspace{-1ex}
\section{Experiments}\vspace{-2ex}
\textbf{Setup.} We evaluate our method on both the synthetic Replica dataset~\citep{straub2019replica} and the real-world ScanNet++ dataset~\citep{yeshwanth2023scannetpp}. 
More implementation details are in the Appendix.

\begin{figure}[t]
  \centering\vspace{-1ex}
     \includegraphics[width=\linewidth]{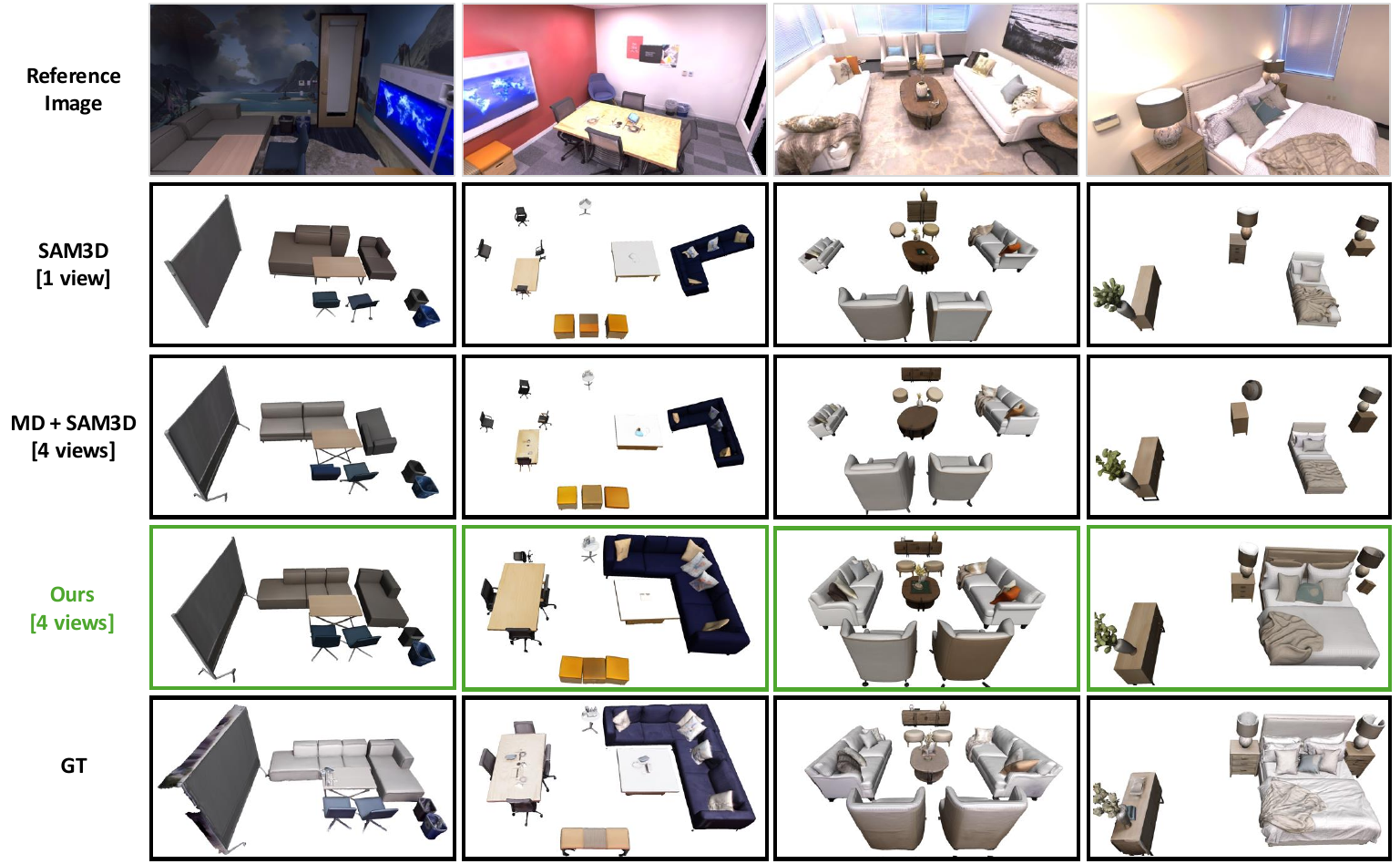}\caption{Qualitative scene-level results on the Replica datasets.}
  \label{fig:scene}
\end{figure}

\begin{table*}[t]\vspace{-3ex}
    \centering
    \caption{\textbf{Scene-level quantitative comparison on Replica.} BP denotes direct backprojection, ICP denotes refinement from the BP using masked depth points, and GS denotes our proposed refinement.}
    \label{tab:comparison_replica}
    \renewcommand{\arraystretch}{1.25}
    \newcolumntype{C}[1]{>{\centering\arraybackslash}p{#1}}
    \resizebox{0.9\textwidth}{!}{
    \begin{tabular}{l l c C{0.75cm} C{0.75cm} C{1.3cm} C{0.75cm} C{0.75cm} C{0.75cm} C{0.75cm}}
        \toprule
        \multicolumn{3}{c}{} &
        \multicolumn{3}{c}{\textit{Geometry}} &
        \multicolumn{4}{c}{\textit{Reconstruction}} \\
        \cmidrule(lr){4-6} \cmidrule(lr){7-10}
        \textbf{Method} &
        \textbf{View} &
        \textbf{Align} &
        CD$^\downarrow$ &
        NC$^\uparrow$ &
        \makecell{F@0.05$^\uparrow$} &
        \makecell{IoU$^\uparrow$} &
        PSNR$^\uparrow$ &
        SSIM$^\uparrow$ &
        LPIPS$^\downarrow$ \\
        \midrule
        $\operatorname{SAM3D}$ \cite{sam3d}     & $\operatorname{1 (random)}$   & $\operatorname{BP}$  & 10.94 & 0.58 & 0.40 & 0.27 & 10.03 & 0.18 & 0.46 \\
        $\operatorname{SAM3D}$ \cite{sam3d}     & $\operatorname{1 (random)}$   & $\operatorname{ICP}$ &  9.49 & 0.62 & 0.52 & 0.29 & 10.03 & 0.19 & 0.45 \\
        \midrule
        $\operatorname{MD}$ \cite{multidiffusion}+$\operatorname{SAM3D}$ \cite{sam3d} & $\operatorname{4 (random)}$   & $\operatorname{BP}$  & 11.01 & 0.58 & 0.39 & 0.27 & 10.19 & 0.19 & 0.46 \\
        $\operatorname{MD}$ \cite{multidiffusion}+$\operatorname{SAM3D}$ \cite{sam3d} & $\operatorname{4 (random)}$   & $\operatorname{ICP}$ &  9.03 & 0.62 & 0.52 & 0.32 & 10.11 & 0.19 & 0.45 \\
        $\operatorname{MD}$ \cite{multidiffusion}+$\operatorname{SAM3D}$ \cite{sam3d} & $\operatorname{4 (coverage)}$ & $\operatorname{BP}$  & 10.72 & 0.59 & 0.42 & 0.46 & 10.55 & 0.24 & 0.46 \\
        $\operatorname{MD}$ \cite{multidiffusion}+$\operatorname{SAM3D}$ \cite{sam3d} & $\operatorname{4 (coverage)}$ & $\operatorname{ICP}$ &  8.83 & 0.63 & 0.54 & 0.49 & 10.60 & 0.24 & 0.45 \\
        \midrule
        $\operatorname{Ours}$ & $\operatorname{4 (coverage)}$ & $\operatorname{BP}$  & 10.24 & 0.59 & 0.41 & 0.48 & 10.67 & 0.25 & 0.46 \\
        $\operatorname{Ours}$ & $\operatorname{4 (coverage)}$ & $\operatorname{ICP}$ &  8.64 & 0.63 & 0.53 & 0.49 & 10.67 & 0.24 & 0.45 \\
        \rowcolor{headerblue}
        $\operatorname{Ours}$ & $\operatorname{4 (coverage)}$ & $\operatorname{GS}$  & \textbf{4.86} & \textbf{0.73} & \textbf{0.73} & \textbf{0.69} & \textbf{12.81} & \textbf{0.38} & \textbf{0.34} \\
        \bottomrule
    \end{tabular}\vspace{-2ex}
    }\vspace{-2ex}
\end{table*}

\vspace{-1ex}
\subsection{Main Results}\vspace{-1ex}
\textbf{Quantitative Results.} 
At the instance level, Table~\ref{tab:main_results} shows that occupancy-coverage view selection improves reconstruction quality in both single-view and multi-view settings.
Replacing random views with coverage-selected views reduces SAM3D CD by 32.9\% on Replica and 27.2\% on ScanNet++.
Interestingly, existing multi-view methods such as InstantMesh and ReconViaGen do not consistently surpass strong completion-oriented single-view models like Amodal3R and SAM3D, highlighting the importance of amodal priors under severe occlusion.
By contrast, our fusion strategy helps SAM3D better exploit complementary views, achieving the best CD, Comp, and F@0.05 on both datasets.
Under the same four-view coverage setting, our method further outperforms MD+SAM3D, reducing CD by 3.9\% and Comp error by 4.4\% on Replica, and reducing CD by 2.0\% while improving F@0.05 by 1.7\% on ScanNet++.
These results show that Scene-SAM3D benefits from both informative view selection and conflict-aware multi-view fusion.

At the scene level, Table~\ref{tab:comparison_replica} and Table~\ref{tab:comparison_scannetpp} compare different layout alignment strategies. 
Under the same four coverage-selected views and fusion strategy, replacing ICP with GS consistently improves both geometry and reconstruction quality. 
Compared with ICP, GS reduces CD by 43.8\%/30.9\%, improves IoU by 40.8\%/42.5\%, and lowers LPIPS by 24.4\%/15.9\% on Replica/ScanNet++, respectively.
These results show that directly backprojecting SAM3D-predicted RST parameters is insufficient for accurate indoor scene layout recovery. 
By contrast, our rendering-based Gaussian layout refinement fully exploits scene-level multi-view cues, including RGB, depth, masks, and other view-level consistency constraints, leading to more coherent scene geometry and higher appearance fidelity.



\textbf{Qualitative Results.}
All results use views selected by our strategy. At the instance level, Fig.~\ref{fig:instance} shows that single-view methods are vulnerable to occlusion and truncation: Trellis and Amodal3R often produce chairs and tables with missing or incomplete structures. Multi-view methods use additional observations but may still suffer from cross-view inconsistencies; ReconViaGen yields distorted reconstructions of the chair and fire extinguisher, while MultiDiffusion introduces redundant structures in the bin and table cases. By contrast, Scene-SAM3D produces more complete geometry while suppressing duplicated, misaligned, and hallucinated structures, and can even recover parts missing or fragmented in the ground-truth mesh.

Fig.~\ref{fig:scene} shows qualitative scene-level comparisons. 
After backprojecting generated objects into the world frame, SAM3D and MD+SAM3D still exhibit noticeable layout errors, even after ICP refinement. 
These include low-quality instances and inaccurate object scales and orientations. 
For example, the office sofa in the second column is undersized, while the chairs in the first column and stools in the third show flipped or unstable orientations. 
Moreover, inaccurate inter-object relationships lead to floating objects or collisions, such as the lamp--cabinet arrangement in the fourth column. 
In contrast, our GS layout refinement produces more plausible RST and spatial relationships.

\subsection{Ablation Study}\vspace{-2ex}

We conduct ablation studies on Replica~\citep{straub2019replica} to analyze the key components of Scene-SAM3D. We present ablations on anchor guidance, fusion steps, and efficiency in the main paper, and provide additional studies on view selection and fusion strategies in Appendix~\ref{sec: more_ablation}.
\begin{figure}[t]\vspace{-3ex}
  \centering
  \begin{minipage}[c]{0.35\linewidth}
    \centering
    \captionof{table}{Ablation of anchor masks.}
    \label{tab:anchor}
    \setlength{\tabcolsep}{6pt}
    \renewcommand{\arraystretch}{1.3}
    \resizebox{\linewidth}{!}{%
    \begin{tabular}{ccccc}
      \toprule
      \textbf{Mask} & \textbf{CD$^\downarrow$} & \textbf{Comp$^\downarrow$} & \textbf{NC$^\uparrow$} & \textbf{F@0.05$^\uparrow$} \\
      \midrule
      $\mathcal{A}$ & 41.37 & 32.39 & 74.85 & 0.72 \\
      $\mathcal{B}$ & 40.38 & 32.13 & 75.24 & 0.72 \\
      \midrule
      \textbf{Ours} & \textbf{39.37} & \textbf{30.65} & \textbf{75.17} & \textbf{0.73} \\
      \bottomrule
    \end{tabular}}

    \vspace{5pt}

    \captionof{table}{Efficiency study.}
    \label{tab:efficiency}
    \setlength{\tabcolsep}{6pt}
    \renewcommand{\arraystretch}{1.25}
    \resizebox{\linewidth}{!}{%
    \begin{tabular}{l c c c}
      \toprule
      \textbf{Method} & \textbf{\# View} & \textbf{TFLOPs} & \textbf{Time (s)} \\
      \midrule
      $\operatorname{MD}$ \cite{multidiffusion}+$\operatorname{SAM3D}$ \cite{sam3d} & 10 & 3,120 & 43.18 \\
      \rowcolor{headerblue}
      \textbf{Ours}  & \textbf{10} & \textbf{2,520}  & \textbf{34.83} \\
      \bottomrule
    \end{tabular}}
  \end{minipage}
  \hfill
  \begin{minipage}[c]{0.55\linewidth}
    \centering
    \includegraphics[width=\linewidth]{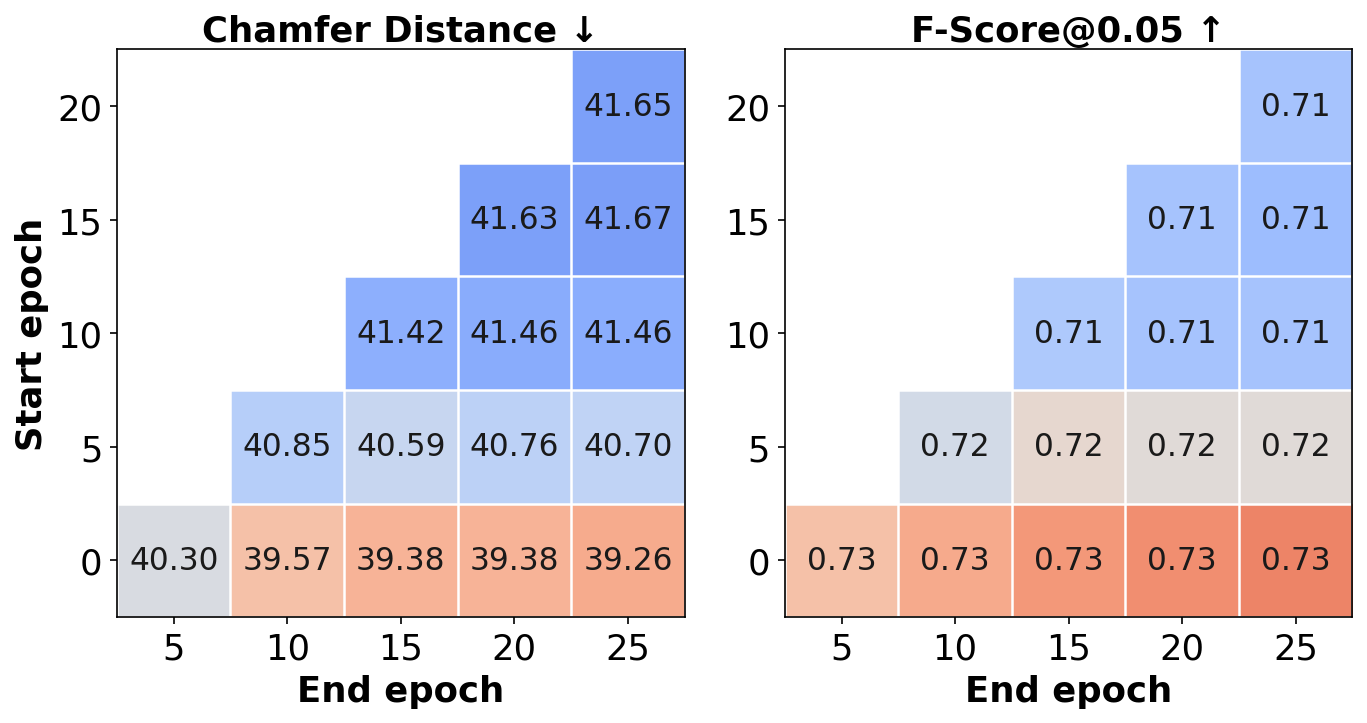}
    \caption{Ablation study on the multiview fusion steps.}
    \label{fig:heatmap}
  \end{minipage}\vspace{-3ex}
\end{figure}

\textbf{Anchor Guidance in Velocity Fusion.}
We further evaluate the privilege of the anchor view during velocity fusion. 
Besides our default design, where the anchor is used as the reference only in its visible region $u \in \Omega_a^{\mathrm{obs}}$, we compare two variants: 
$\mathcal{A}$ removes anchor privilege and lets all selected views vote at every position, while 
$\mathcal{B}$ assigns anchor privilege to all positions, including $u \in \Omega_a^{\mathrm{unobs}}$. 
As shown in Tab.~\ref{tab:anchor}, using the anchor as reference only in its observed region achieves the best performance.
This strategy leverages the reliable local evidence from the anchor while preventing hallucinated anchor predictions in unobserved regions from being propagated to the final geometry. 


\textbf{Impact of Fusion Steps.}
We evaluate how the fusion start and end steps affect generation quality.
As shown in Fig.~\ref{fig:heatmap}, warmer colors indicate better quality.
Starting fusion from step 0 achieves the best results, suggesting that multi-view cues are most useful during early structure formation and geometry completion.
In contrast, continuing fusion from the middle steps to the end brings only marginal gains and may introduce canonical-space conflicts from view-specific refinements.
We therefore adopt early-stop fusion to reduce computational cost while maintaining geometric quality.


\textbf{Efficiency Study in Multiview Fusion.}
We further quantitatively evaluate the efficiency gains brought by our step-efficient fusion strategy.
Since the remaining components of the pipeline are kept unchanged, we focus on comparing the flow-model inference time and FLOPs against the MultiDiffusion fusion baseline.
The main cost comes from repeated forward evaluations of the flow velocity model: MultiDiffusion fusion runs the full flow process for every selected view at every inference step, whereas our early-stop fusion only uses helper-view conditions in the early steps and switches to anchor-only inference afterwards.
As shown in Table~\ref{tab:efficiency}, under ten-view inference, our method reduces both flow-model FLOPs and flow-model sampling wall-time latency by nearly 20\%.

\vspace{-2ex}
\section{Conclusion and Future Work}\vspace{-2ex}

We introduced Scene-SAM3D, a training-free framework that extends object-centric single-view 3D foundation models to calibrated multi-view scene asset generation. By combining Occupancy-Coverage View Selection, Step-Efficient Conflict-Aware Multi-view Fusion, and Rigid-Object Gaussian Layout Refinement, it addresses redundant observations, inconsistent multi-view evidence, and incoherent scene layouts. Several challenges remain. Current object-centric models such as SAM3D and TRELLIS require separate inference for each instance, limiting scene-level scalability. Moreover, our view selection and fusion strategy is shared across both SAM3D stages and remains primarily geometry-oriented. Future work will explore joint multi-instance generation and stage-aware strategies that use different view evidence for geometry and texture.

\clearpage
\bibliographystyle{unsrtnat}
\bibliography{references}

@inproceedings{wang2018pixel2mesh,
  author    = {Nanyang Wang and Yinda Zhang and Zhuwen Li and Yanwei Fu and Wei Liu and Yu{-}Gang Jiang},
  title     = {Pixel2Mesh: Generating 3D Mesh Models from Single {RGB} Images},
  booktitle = {Computer Vision - {ECCV} 2018},
  pages     = {55--71},
  publisher = {Springer},
  year      = {2018},
}

@article{liu2019softras,
  author  = {Shichen Liu and Weikai Chen and Tianye Li and Hao Li},
  title   = {Soft Rasterizer: Differentiable Rendering for Unsupervised Single-View Mesh Reconstruction},
  journal = {CoRR},
  volume  = {abs/1901.05567},
  year    = {2019},
}

@inproceedings{kato2018nmr,
  author    = {Hiroharu Kato and Yoshitaka Ushiku and Tatsuya Harada},
  title     = {Neural 3D Mesh Renderer},
  booktitle = {IEEE/CVF Conference on Computer Vision and Pattern Recognition ({CVPR})},
  pages     = {3907--3916},
  year      = {2018},
}

@inproceedings{xu2019disn,
  author    = {Qiangeng Xu and Weiyue Wang and Duygu Ceylan and Radom{\'{\i}}r Mech and Ulrich Neumann},
  title     = {{DISN:} Deep Implicit Surface Network for High-quality Single-view 3D Reconstruction},
  booktitle = {Advances in Neural Information Processing Systems ({NeurIPS})},
  pages     = {490--500},
  year      = {2019},
}

@inproceedings{hong2024lrm,
  author    = {Yicong Hong and Kai Zhang and Jiuxiang Gu and Sai Bi and Yang Zhou and Difan Liu and Feng Liu and Kalyan Sunkavalli and Trung Bui and Hao Tan},
  title     = {{LRM:} Large Reconstruction Model for Single Image to 3D},
  booktitle = {The Twelfth International Conference on Learning Representations ({ICLR})},
  publisher = {OpenReview.net},
  year      = {2024},
}

@inproceedings{wu2024direct3d,
  author    = {Shuang Wu and Youtian Lin and Yifei Zeng and Feihu Zhang and Jingxi Xu and Philip Torr and Xun Cao and Yao Yao},
  title     = {Direct3D: Scalable Image-to-3D Generation via 3D Latent Diffusion Transformer},
  booktitle = {Advances in Neural Information Processing Systems ({NeurIPS})},
  year      = {2024},
}

@inproceedings{liu2023one2345,
  author    = {Minghua Liu and Chao Xu and Haian Jin and Linghao Chen and Mukund Varma T. and Zexiang Xu and Hao Su},
  title     = {One-2-3-45: Any Single Image to 3D Mesh in 45 Seconds without Per-Shape Optimization},
  booktitle = {Advances in Neural Information Processing Systems ({NeurIPS})},
  year      = {2023},
}

@inproceedings{wang2024crm,
  author    = {Zhengyi Wang and Yikai Wang and Yifei Chen and Chendong Xiang and Shuo Chen and Dajiang Yu and Chongxuan Li and Hang Su and Jun Zhu},
  title     = {{CRM:} Single Image to 3D Textured Mesh with Convolutional Reconstruction Model},
  booktitle = {Computer Vision - {ECCV} 2024},
  pages     = {57--74},
  publisher = {Springer},
  year      = {2024},
}

@inproceedings{long2024wonder3d,
  author    = {Xiaoxiao Long and Yuan{-}Chen Guo and Cheng Lin and Yuan Liu and Zhiyang Dou and Lingjie Liu and Yuexin Ma and Song{-}Hai Zhang and Marc Habermann and Christian Theobalt and Wenping Wang},
  title     = {Wonder3D: Single Image to 3D Using Cross-Domain Diffusion},
  booktitle = {{IEEE/CVF} Conference on Computer Vision and Pattern Recognition ({CVPR})},
  pages     = {9970--9980},
  publisher = {{IEEE}},
  year      = {2024},
}

@article{xu2024instantmesh,
  author  = {Jiale Xu and Weihao Cheng and Yiming Gao and Xintao Wang and Shenghua Gao and Ying Shan},
  title   = {InstantMesh: Efficient 3D Mesh Generation from a Single Image with Sparse-view Large Reconstruction Models},
  journal = {CoRR},
  volume  = {abs/2404.07191},
  year    = {2024},
}

@inproceedings{wu2024unique3d,
  author    = {Kailu Wu and Fangfu Liu and Zhihan Cai and Runjie Yan and Hanyang Wang and Yating Hu and Yueqi Duan and Kaisheng Ma},
  title     = {Unique3D: High-Quality and Efficient 3D Mesh Generation from a Single Image},
  booktitle = {Advances in Neural Information Processing Systems ({NeurIPS})},
  year      = {2024},
}

@inproceedings{trellis,
  author    = {Jianfeng Xiang and Zelong Lv and Sicheng Xu and Yu Deng and Ruicheng Wang and Bowen Zhang and Dong Chen and Xin Tong and Jiaolong Yang},
  title     = {Structured 3D Latents for Scalable and Versatile 3D Generation},
  booktitle = {{IEEE/CVF} Conference on Computer Vision and Pattern Recognition ({CVPR})},
  pages     = {21469--21480},
  year      = {2025},
}

@article{zhao2025hunyuan3d,
  author  = {Zibo Zhao and Zeqiang Lai and Qingxiang Lin and Yunfei Zhao and Haolin Liu and Shuhui Yang and Yifei Feng and Mingxin Yang and Sheng Zhang and Xianghui Yang and Huiwen Shi and Sicong Liu and Junta Wu and Yihang Lian and Fan Yang and Ruining Tang and Zebin He and Xinzhou Wang and Jian Liu and Xuhui Zuo and Zhuo Chen and Biwen Lei and Haohan Weng and Jing Xu and Yiling Zhu and Xinhai Liu and Lixin Xu and Changrong Hu and Tianyu Huang and Lifu Wang and Jihong Zhang and Meng Chen and Liang Dong and Yiwen Jia and Yulin Cai and Jiaao Yu and Yixuan Tang and Hao Zhang and Zheng Ye and Peng He and Runzhou Wu and Chao Zhang and Yonghao Tan and Jie Xiao and Yangyu Tao and Jianchen Zhu and Jinbao Xue and Kai Liu and Chongqing Zhao and Xinming Wu and Zhichao Hu and Lei Qin and Jianbing Peng and Zhan Li and Minghui Chen and Xipeng Zhang and Lin Niu and Paige Wang and Yingkai Wang and Haozhao Kuang and Zhongyi Fan and Xu Zheng and Weihao Zhuang and YingPing He and Tian Liu and Yong Yang and Di Wang and Yuhong Liu and Jie Jiang and Jingwei Huang and Chunchao Guo},
  title   = {Hunyuan3D 2.0: Scaling Diffusion Models for High Resolution Textured 3D Assets Generation},
  journal = {CoRR},
  volume  = {abs/2501.12202},
  year    = {2025},
}

@article{li2025triposg,
  author    = {Yangguang Li and Zi-Xin Zou and Zexiang Liu and Dehu Wang and Yuan Liang and Zhipeng Yu and Xingchao Liu and Yuan-Chen Guo and Ding Liang and Wanli Ouyang and others},
  title     = {{TriposG:} High-Fidelity 3D Shape Synthesis Using Large-Scale Rectified Flow Models},
  journal   = {{IEEE} Transactions on Pattern Analysis and Machine Intelligence},
  year      = {2025},
  publisher = {{IEEE}},
}

@inproceedings{huang2025midi,
  author    = {Zehuan Huang and Yuan{-}Chen Guo and Xingqiao An and Yunhan Yang and Yangguang Li and Zi{-}Xin Zou and Ding Liang and Xihui Liu and Yan{-}Pei Cao and Lu Sheng},
  title     = {{MIDI:} Multi-Instance Diffusion for Single Image to 3D Scene Generation},
  booktitle = {{IEEE/CVF} Conference on Computer Vision and Pattern Recognition ({CVPR})},
  pages     = {23646--23657},
  publisher = {Computer Vision Foundation / {IEEE}},
  year      = {2025},
}

@article{meng2025scenegen,
  author  = {Yanxu Meng and Haoning Wu and Ya Zhang and Weidi Xie},
  title   = {SceneGen: Single-Image 3D Scene Generation in One Feedforward Pass},
  journal = {CoRR},
  volume  = {abs/2508.15769},
  year    = {2025},
}

@article{shi2025scenemaker,
  author  = {Yukai Shi and Weiyu Li and Zihao Wang and Hongyang Li and Xingyu Chen and Ping Tan and Lei Zhang},
  title   = {SceneMaker: Open-set 3D Scene Generation with Decoupled De-occlusion and Pose Estimation Model},
  journal = {CoRR},
  volume  = {abs/2512.10957},
  year    = {2025},
}

@article{ling2025iscene,
  author  = {Lu Ling and Yunhao Ge and Yichen Sheng and Aniket Bera},
  title   = {I-Scene: 3D Instance Models are Implicit Generalizable Spatial Learners},
  journal = {CoRR},
  volume  = {abs/2512.13683},
  year    = {2025},
}

@inproceedings{wu2025amodal3r,
  author    = {Tianhao Wu and Chuanxia Zheng and Frank Guan and Andrea Vedaldi and Tat{-}Jen Cham},
  title     = {Amodal3R: Amodal 3D Reconstruction from Occluded 2D Images},
  booktitle = {Proceedings of the {IEEE/CVF} International Conference on Computer Vision ({ICCV})},
  pages     = {9181--9193},
  year      = {2025},
}

@article{sam3d,
  author  = {Xingyu Chen and Fu{-}Jen Chu and Pierre Gleize and Kevin J. Liang and Alexander Sax and Hao Tang and Weiyao Wang and Michelle Guo and Thibaut Hardin and Xiang Li and Aohan Lin and Jiawei Liu and Ziqi Ma and Anushka Sagar and Bowen Song and Xiaodong Wang and Jianing Yang and Bowen Zhang and Piotr Doll{\'{a}}r and Georgia Gkioxari and Matt Feiszli and Jitendra Malik},
  title   = {{SAM} 3D: 3Dfy Anything in Images},
  journal = {CoRR},
  volume  = {abs/2511.16624},
  year    = {2025},
}

@inproceedings{multidiffusion,
  author       = {Omer Bar{-}Tal and
                  Lior Yariv and
                  Yaron Lipman and
                  Tali Dekel},
  editor       = {Andreas Krause and
                  Emma Brunskill and
                  Kyunghyun Cho and
                  Barbara Engelhardt and
                  Sivan Sabato and
                  Jonathan Scarlett},
  title        = {MultiDiffusion: Fusing Diffusion Paths for Controlled Image Generation},
  booktitle    = {International Conference on Machine Learning, {ICML}},
  pages        = {1737--1752},
  year         = {2023},
}

@inproceedings{liu2024syncdreamer,
  author    = {Yuan Liu and Cheng Lin and Zijiao Zeng and Xiaoxiao Long and Lingjie Liu and Taku Komura and Wenping Wang},
  title     = {SyncDreamer: Generating Multiview-consistent Images from a Single-view Image},
  booktitle = {The Twelfth International Conference on Learning Representations ({ICLR})},
  publisher = {OpenReview.net},
  year      = {2024},
}

@inproceedings{voleti2024sv3d,
  author    = {Vikram Voleti and Chun{-}Han Yao and Mark Boss and Adam Letts and David Pankratz and Dmitry Tochilkin and Christian Laforte and Robin Rombach and Varun Jampani},
  title     = {{SV3D:} Novel Multi-view Synthesis and 3D Generation from a Single Image Using Latent Video Diffusion},
  booktitle = {Computer Vision - {ECCV} 2024},
  pages     = {439--457},
  publisher = {Springer},
  year      = {2024},
}

@inproceedings{wen2025ouroboros3d,
  author    = {Hao Wen and Zehuan Huang and Yaohui Wang and Xinyuan Chen and Lu Sheng},
  title     = {Ouroboros3D: Image-to-3D Generation via 3D-aware Recursive Diffusion},
  booktitle = {{IEEE/CVF} Conference on Computer Vision and Pattern Recognition ({CVPR})},
  pages     = {21631--21641},
  publisher = {Computer Vision Foundation / {IEEE}},
  year      = {2025},
}

@article{huang2026unirecgen,
  author  = {Zhisheng Huang and Jiahao Chen and Cheng Lin and Chenyu Hu and Hanzhuo Huang and Zhengming Yu and Mengfei Li and Yuheng Liu and Zekai Gu and Zibo Zhao and others},
  title   = {UniRecGen: Unifying Multi-View 3D Reconstruction and Generation},
  journal = {arXiv preprint arXiv:2604.01479},
  year    = {2026},
}

@article{siddiqui2026shaper,
  author  = {Yawar Siddiqui and Duncan P. Frost and Samir Aroudj and Armen Avetisyan and Henry Howard{-}Jenkins and Daniel DeTone and Pierre Moulon and Qirui Wu and Zhengqin Li and Julian Straub and Richard A. Newcombe and Jakob J. Engel},
  title   = {ShapeR: Robust Conditional 3D Shape Generation from Casual Captures},
  journal = {CoRR},
  volume  = {abs/2601.11514},
  year    = {2026},
}

@article{mvsam3d,
  author  = {Baicheng Li and Dong Wu and Jun Li and Shunkai Zhou and Zecui Zeng and Lusong Li and Hongbin Zha},
  title   = {{MV-SAM3D:} Adaptive Multi-View Fusion for Layout-Aware 3D Generation},
  journal = {CoRR},
  volume  = {abs/2603.11633},
  year    = {2026},
}

@article{straub2019replica,
  author  = {Julian Straub and Thomas Whelan and Lingni Ma and Yufan Chen and Erik Wijmans and Simon Green and Jakob J. Engel and Raul Mur{-}Artal and Carl Yuheng Ren and Shobhit Verma and Anton Clarkson and Mingfei Yan and Brian Budge and Yajie Yan and Xiaqing Pan and June Yon and Yuyang Zou and Kimberly Leon and Nigel Carter and Jesus Briales and Tyler Gillingham and Elias Mueggler and Luis Pesqueira and Manolis Savva and Dhruv Batra and Hauke M. Strasdat and Renzo {De Nardi} and Michael Goesele and Steven Lovegrove and Richard A. Newcombe},
  title   = {The Replica Dataset: {A} Digital Replica of Indoor Spaces},
  journal = {CoRR},
  volume  = {abs/1906.05797},
  year    = {2019},
}

@inproceedings{yeshwanth2023scannetpp,
  author    = {Chandan Yeshwanth and Yueh{-}Cheng Liu and Matthias Nie{\ss}ner and Angela Dai},
  title     = {ScanNet++: {A} High-Fidelity Dataset of 3D Indoor Scenes},
  booktitle = {{IEEE/CVF} International Conference on Computer Vision ({ICCV})},
  pages     = {12--22},
  publisher = {{IEEE}},
  year      = {2023},
}

@inproceedings{ni2025dprecon,
  author    = {Junfeng Ni and Yu Liu and Ruijie Lu and Zirui Zhou and Song{-}Chun Zhu and Yixin Chen and Siyuan Huang},
  title     = {Decompositional Neural Scene Reconstruction with Generative Diffusion Prior},
  booktitle = {{IEEE/CVF} Conference on Computer Vision and Pattern Recognition ({CVPR})},
  pages     = {6022--6033},
  publisher = {Computer Vision Foundation / {IEEE}},
  year      = {2025},
}

@inproceedings{kong2023vmap,
  author    = {Xin Kong and Shikun Liu and Marwan Taher and Andrew J. Davison},
  title     = {vMAP: Vectorised Object Mapping for Neural Field {SLAM}},
  booktitle = {{IEEE/CVF} Conference on Computer Vision and Pattern Recognition ({CVPR})},
  pages     = {952--961},
  publisher = {{IEEE}},
  year      = {2023},
}

@inproceedings{wang2024robogen,
  author    = {Yufei Wang and Zhou Xian and Feng Chen and Tsun{-}Hsuan Wang and Yian Wang and Katerina Fragkiadaki and Zackory Erickson and David Held and Chuang Gan},
  title     = {RoboGen: Towards Unleashing Infinite Data for Automated Robot Learning via Generative Simulation},
  booktitle = {Forty-first International Conference on Machine Learning ({ICML})},
  series    = {Proceedings of Machine Learning Research},
  pages     = {51936--51983},
  publisher = {{PMLR} / OpenReview.net},
  year      = {2024},
}

@inproceedings{gu2023maniskill2,
  author    = {Jiayuan Gu and Fanbo Xiang and Xuanlin Li and Zhan Ling and Xiqiang Liu and Tongzhou Mu and Yihe Tang and Stone Tao and Xinyue Wei and Yunchao Yao and Xiaodi Yuan and Pengwei Xie and Zhiao Huang and Rui Chen and Hao Su},
  title     = {ManiSkill2: A Unified Benchmark for Generalizable Manipulation Skills},
  booktitle = {The Eleventh International Conference on Learning Representations ({ICLR})},
  publisher = {OpenReview.net},
  year      = {2023},
}

@inproceedings{khanna2024hssd200,
  author    = {Mukul Khanna and Yongsen Mao and Hanxiao Jiang and Sanjay Haresh and Brennan Shacklett and Dhruv Batra and Alexander Clegg and Eric Undersander and Angel X. Chang and Manolis Savva},
  title     = {Habitat Synthetic Scenes Dataset ({HSSD-200}): An Analysis of 3D Scene Scale and Realism Tradeoffs for ObjectGoal Navigation},
  booktitle = {{IEEE/CVF} Conference on Computer Vision and Pattern Recognition ({CVPR})},
  pages     = {16384--16393},
  publisher = {{IEEE}},
  year      = {2024},
}

@inproceedings{puig2024habitat3,
  author    = {Xavier Puig and Eric Undersander and Andrew Szot and Mikael Dallaire Cote and Tsung{-}Yen Yang and Ruslan Partsey and Ruta Desai and Alexander Clegg and Michal Hlavac and So Yeon Min and Vladimir Vondrus and Th{\'{e}}ophile Gervet and Vincent{-}Pierre Berges and John M. Turner and Oleksandr Maksymets and Zsolt Kira and Mrinal Kalakrishnan and Jitendra Malik and Devendra Singh Chaplot and Unnat Jain and Dhruv Batra and Akshara Rai and Roozbeh Mottaghi},
  title     = {Habitat 3.0: A Co-Habitat for Humans, Avatars, and Robots},
  booktitle = {The Twelfth International Conference on Learning Representations ({ICLR})},
  publisher = {OpenReview.net},
  year      = {2024},
}

@inproceedings{jones2014roomalive,
  author    = {Brett R. Jones and Rajinder Sodhi and Michael Murdock and Ravish Mehra and Hrvoje Benko and Andrew D. Wilson and Eyal Ofek and Blair MacIntyre and Nikunj Raghuvanshi and Lior Shapira},
  title     = {RoomAlive: Magical Experiences Enabled by Scalable, Adaptive Projector-Camera Units},
  booktitle = {The 27th Annual {ACM} Symposium on User Interface Software and Technology ({UIST})},
  pages     = {637--644},
  publisher = {{ACM}},
  year      = {2014},
}

@inproceedings{orts2016holoportation,
  author    = {Sergio Orts{-}Escolano and Christoph Rhemann and Sean Ryan Fanello and Wayne Chang and Adarsh Kowdle and Yury Degtyarev and David Kim and Philip Davidson and Sameh Khamis and Mingsong Dou and Vladimir Tankovich and Charles T. Loop and Qin Cai and Philip A. Chou and Sarah Mennicken and Julien P. C. Valentin and Vivek Pradeep and Shenlong Wang and Sing Bing Kang and Pushmeet Kohli and Yuliya Lutchyn and Cem Keskin and Shahram Izadi},
  title     = {Holoportation: Virtual 3D Teleportation in Real-Time},
  booktitle = {Proceedings of the 29th Annual Symposium on User Interface Software and Technology ({UIST})},
  pages     = {741--754},
  publisher = {{ACM}},
  year      = {2016},
}

@inproceedings{newcombe2011kinectfusion,
  author    = {Richard A. Newcombe and Shahram Izadi and Otmar Hilliges and David Molyneaux and David Kim and Andrew J. Davison and Pushmeet Kohli and Jamie Shotton and Steve Hodges and Andrew Fitzgibbon},
  title     = {KinectFusion: Real-Time Dense Surface Mapping and Tracking},
  booktitle = {10th {IEEE} International Symposium on Mixed and Augmented Reality ({ISMAR})},
  pages     = {127--136},
  publisher = {{IEEE}},
  year      = {2011},
}

@article{sra2018oasis,
  author  = {Misha Sra and Sergio Garrido{-}Jurado and Pattie Maes},
  title   = {Oasis: Procedurally Generated Social Virtual Spaces from 3D Scanned Real Spaces},
  journal = {{IEEE} Trans. Vis. Comput. Graph.},
  volume  = {24},
  number  = {12},
  pages   = {3174--3187},
  year    = {2018},
}

@inproceedings{xu2023vrnerf,
  author    = {Linning Xu and Vasu Agrawal and William Laney and Tony Garcia and Aayush Bansal and Changil Kim and Samuel Rota Bul{\`o} and Lorenzo Porzi and Peter Kontschieder and Alja{\v{z}} Bo{\v{z}}i{\v{c}} and others},
  title     = {{VR-NeRF:} High-Fidelity Virtualized Walkable Spaces},
  booktitle = {{SIGGRAPH} Asia Conference Papers},
  pages     = {1--12},
  year      = {2023},
}

@article{huang2026litereality,
  author  = {Zhening Huang and Xiaoyang Wu and Fangcheng Zhong and Hengshuang Zhao and Matthias Nie{\ss}ner and Joan Lasenby},
  title   = {LiteReality: Graphics-Ready 3D Scene Reconstruction from {RGB-D} Scans},
  journal = {Advances in Neural Information Processing Systems},
  volume  = {38},
  pages   = {162794--162827},
  year    = {2026},
}

@inproceedings{ma2023sqa3d,
  author    = {Xiaojian Ma and Silong Yong and Zilong Zheng and Qing Li and Yitao Liang and Song{-}Chun Zhu and Siyuan Huang},
  title     = {{SQA3D:} Situated Question Answering in 3D Scenes},
  booktitle = {The Eleventh International Conference on Learning Representations ({ICLR})},
  publisher = {OpenReview.net},
  year      = {2023},
}

@inproceedings{majumdar2024openeqa,
  author    = {Arjun Majumdar and Anurag Ajay and Xiaohan Zhang and Pranav Putta and Sriram Yenamandra and Mikael Henaff and Sneha Silwal and Paul McVay and Oleksandr Maksymets and Sergio Arnaud and Karmesh Yadav and Qiyang Li and Ben Newman and Mohit Sharma and Vincent{-}Pierre Berges and Shiqi Zhang and Pulkit Agrawal and Yonatan Bisk and Dhruv Batra and Mrinal Kalakrishnan and Franziska Meier and Chris Paxton and Alexander Sax and Aravind Rajeswaran},
  title     = {OpenEQA: Embodied Question Answering in the Era of Foundation Models},
  booktitle = {{IEEE/CVF} Conference on Computer Vision and Pattern Recognition ({CVPR})},
  pages     = {16488--16498},
  publisher = {{IEEE}},
  year      = {2024},
}

@inproceedings{deitke2023objaversexl,
  author    = {Matt Deitke and Ruoshi Liu and Matthew Wallingford and Huong Ngo and Oscar Michel and Aditya Kusupati and Alan Fan and Christian Laforte and Vikram Voleti and Samir Yitzhak Gadre and Eli VanderBilt and Aniruddha Kembhavi and Carl Vondrick and Georgia Gkioxari and Kiana Ehsani and Ludwig Schmidt and Ali Farhadi},
  title     = {Objaverse-XL: A Universe of 10M+ 3D Objects},
  booktitle = {Advances in Neural Information Processing Systems ({NeurIPS})},
  year      = {2023},
}

@inproceedings{collins2022abo,
  author    = {Jasmine Collins and Shubham Goel and Kenan Deng and Achleshwar Luthra and Leon Xu and Erhan Gundogdu and Xi Zhang and Tomas F. Yago Vicente and Thomas Dideriksen and Himanshu Arora and Matthieu Guillaumin and Jitendra Malik},
  title     = {{ABO:} Dataset and Benchmarks for Real-World 3D Object Understanding},
  booktitle = {{IEEE/CVF} Conference on Computer Vision and Pattern Recognition ({CVPR})},
  pages     = {21094--21104},
  publisher = {{IEEE}},
  year      = {2022},
}

@inproceedings{liu2023zero1to3,
  author    = {Ruoshi Liu and Rundi Wu and Basile Van Hoorick and Pavel Tokmakov and Sergey Zakharov and Carl Vondrick},
  title     = {Zero-1-to-3: Zero-shot One Image to 3D Object},
  booktitle = {{IEEE/CVF} International Conference on Computer Vision ({ICCV})},
  pages     = {9264--9275},
  publisher = {{IEEE}},
  year      = {2023},
}

@inproceedings{han2025flex3d,
  author    = {Junlin Han and Jianyuan Wang and Andrea Vedaldi and Philip Torr and Filippos Kokkinos},
  title     = {Flex3D: Feed-Forward 3D Generation with Flexible Reconstruction Model and Input View Curation},
  booktitle = {Forty-second International Conference on Machine Learning ({ICML})},
  series    = {Proceedings of Machine Learning Research},
  publisher = {{PMLR} / OpenReview.net},
  year      = {2025},
}

@article{pito1999nextbestview,
  author  = {Richard Pito},
  title   = {A Solution to the Next Best View Problem for Automated Surface Acquisition},
  journal = {{IEEE} Trans. Pattern Anal. Mach. Intell.},
  volume  = {21},
  number  = {10},
  pages   = {1016--1030},
  year    = {1999},
}

@inproceedings{border2018see,
  author    = {Rowan Border and Jonathan D. Gammell and Paul Newman},
  title     = {Surface Edge Explorer (SEE): Planning Next Best Views Directly from 3D Observations},
  booktitle = {2018 {IEEE} International Conference on Robotics and Automation ({ICRA})},
  pages     = {1--8},
  publisher = {{IEEE}},
  year      = {2018},
}

@article{border2024see,
  author  = {Rowan Border and Jonathan D. Gammell},
  title   = {The Surface Edge Explorer (SEE): A Measurement-Direct Approach to Next Best View Planning},
  journal = {Int. J. Robotics Res.},
  volume  = {43},
  number  = {10},
  pages   = {1506--1532},
  year    = {2024},
}

@inproceedings{jiang2024fisherrf,
  author    = {Wen Jiang and Boshu Lei and Kostas Daniilidis},
  title     = {FisherRF: Active View Selection and Mapping with Radiance Fields Using Fisher Information},
  booktitle = {Computer Vision - {ECCV} 2024},
  pages     = {422--440},
  publisher = {Springer},
  year      = {2024},
}

@inproceedings{wilson2025popgs,
  author    = {Joey Wilson and Marcelino Almeida and Sachit Mahajan and Martin Labrie and Maani Ghaffari and Omid Ghasemalizadeh and Min Sun and Cheng{-}Hao Kuo and Arnab Sen},
  title     = {POp-GS: Next Best View in 3D-Gaussian Splatting with P-Optimality},
  booktitle = {{IEEE/CVF} Conference on Computer Vision and Pattern Recognition ({CVPR})},
  pages     = {3646--3655},
  publisher = {Computer Vision Foundation / {IEEE}},
  year      = {2025},
}

@inproceedings{peralta2020nbvpolicy,
  author    = {Daryl Peralta and Joel Casimiro and Aldrin Michael Nilles and Justine Aletta Aguilar and Rowel Atienza and Rhandley Cajote},
  title     = {Next-Best View Policy for 3D Reconstruction},
  booktitle = {Computer Vision - {ECCV} 2020 Workshops},
  pages     = {558--573},
  publisher = {Springer},
  year      = {2020},
}

@inproceedings{chen2024gennbv,
  author    = {Xiao Chen and Quanyi Li and Tai Wang and Tianfan Xue and Jiangmiao Pang},
  title     = {GenNBV: Generalizable Next-Best-View Policy for Active 3D Reconstruction},
  booktitle = {{IEEE/CVF} Conference on Computer Vision and Pattern Recognition ({CVPR})},
  pages     = {16436--16445},
  publisher = {{IEEE}},
  year      = {2024},
}

@article{frahm2025vinnbv,
  author     = {Noah Frahm and Dongxu Zhao and Andrea Dunn Beltran and Ron Alterovitz and Jan{-}Michael Frahm and Junier Oliva and Roni Sengupta},
  title      = {{VIN-NBV:} A View Introspection Network for Next-Best-View Selection for Resource-Efficient 3D Reconstruction},
  journal    = {CoRR},
  volume     = {abs/2505.06219},
  year       = {2025},
  eprinttype = {arXiv},
  eprint     = {2505.06219},
}

@article{reconviagen,
  author       = {Jiahao Chang and
                  Chongjie Ye and
                  Yushuang Wu and
                  Yuantao Chen and
                  Yidan Zhang and
                  Zhongjin Luo and
                  Chenghong Li and
                  Yihao Zhi and
                  Xiaoguang Han},
  title        = {ReconViaGen: Towards Accurate Multi-view 3D Object Reconstruction
                  via Generation},
  journal      = {CoRR},
  volume       = {abs/2510.23306},
  year         = {2025},
  eprinttype   = {arXiv},
  eprint       = {2510.23306},
}

@inproceedings{vggt,
  author       = {Jianyuan Wang and
                  Minghao Chen and
                  Nikita Karaev and
                  Andrea Vedaldi and
                  Christian Rupprecht and
                  David Novotn{\'{y}}},
  title        = {{VGGT:} Visual Geometry Grounded Transformer},
  booktitle    = {{IEEE/CVF} Conference on Computer Vision and Pattern Recognition,
                  {CVPR} 2025, Nashville, TN, USA, June 11-15, 2025},
  pages        = {5294--5306},
  publisher    = {Computer Vision Foundation / {IEEE}},
  year         = {2025},
}

@article{icp,
  author       = {Paul J. Besl and
                  Neil D. McKay},
  title        = {A Method for Registration of 3-D Shapes},
  journal      = {{IEEE} Trans. Pattern Anal. Mach. Intell.},
  volume       = {14},
  number       = {2},
  pages        = {239--256},
  year         = {1992},
}

@inproceedings{kmeans++,
  author       = {David Arthur and
                  Sergei Vassilvitskii},
  editor       = {Nikhil Bansal and
                  Kirk Pruhs and
                  Clifford Stein},
  title        = {k-means++: the advantages of careful seeding},
  booktitle    = {Proceedings of the Eighteenth Annual {ACM-SIAM} Symposium on Discrete
                  Algorithms, {SODA} 2007, New Orleans, Louisiana, USA, January 7-9,
                  2007},
  pages        = {1027--1035},
  publisher    = {{SIAM}},
  year         = {2007},
}

\clearpage
\appendix
\section*{Appendix}
\section{Related Work}\label{app:related}
\textbf{Single-image 3D Asset Generation.}
Early single-image 3D reconstruction methods relied on explicit geometric priors, differentiable rendering, or implicit surface modeling~\citep{wang2018pixel2mesh, liu2019softras, kato2018nmr, xu2019disn}. 
Recent progress has shifted toward large generative models that either directly infer 3D representations~\citep{hong2024lrm, wu2024direct3d} or synthesize intermediate views before reconstruction~\citep{liu2023one2345, wang2024crm, long2024wonder3d, xu2024instantmesh, wu2024unique3d}. 
Large-scale 3D asset datasets have further enabled flow-based 3D foundation models with stronger object-level priors~\citep{deitke2023objaversexl, collins2022abo, trellis, zhao2025hunyuan3d, li2025triposg}. 
Recent works extend these priors to compositional or occlusion-aware settings~\citep{huang2025midi, meng2025scenegen, shi2025scenemaker, ling2025iscene, wu2025amodal3r}. 
Most notably, SAM3D~\citep{sam3d} marks a recent leap in natural-scene object reconstruction by jointly predicting shape, texture, and pose from arbitrary single image. 
We build on this strong prior, but move beyond its single-view formulation to multi-view scene asset generation, where uneven view quality, severe occlusion, and cross-view canonical conflicts must be handled explicitly.

\textbf{Multi-view Fusion for 3D Generation.} A common strategy in image-to-3D generation is to synthesize a small set of predefined views and reconstruct 3D geometry from them~\citep{wang2024crm, xu2024instantmesh, wu2024unique3d}. 
Other methods improve cross-view consistency through feature alignment, geometry-aware fusion, video priors, or reconstruction-in-the-loop feedback~\citep{liu2024syncdreamer, long2024wonder3d, voleti2024sv3d, wen2025ouroboros3d}. 
More recent approaches learn genuinely multi-view-conditioned 3D generators from large-scale 3D data~\citep{huang2026unirecgen, siddiqui2026shaper}. Hunyuan3D 2.0~\citep{zhao2025hunyuan3d} also provides a multi-view variant that conditions shape generation on up to four predefined canonical view slots---front, back, left, and right. However, it does not model arbitrary calibrated camera viewpoints, limiting its applicability to irregular multi-view observations captured in real scenes.
ReconViaGen~\citep{reconviagen} further integrates VGGT~\citep{vggt} reconstruction priors into TRELLIS, using multi-view features as global geometry conditions for coarse structure generation and local per-view conditions for fine geometry and texture generation.
Concurrent with our work, MV-SAM3D~\citep{mvsam3d} investigates training-free multi-view fusion in the latent space of a pre-trained SAM3D model, primarily for object-centric inputs. In contrast, Scene-SAM3D targets indoor scene asset generation, where object instances are heavily occluded, observations are highly redundant, and view selection and fusion must handle inconsistent multi-view evidence.

\textbf{View Selection for 3D Generation and Reconstruction.} In generative 3D, many methods still rely on a small set of fixed or pre-defined views, and explicit view selection remains largely underexplored. One-2-3-45~\citep{liu2023one2345} uses a two-stage view generation strategy, where Zero-1-to-3~\citep{liu2023zero1to3} expands each base view into several nearby views and a ground-truth target view is randomly sampled for supervision. This is closer to target-view sampling than explicit input-view selection.
Flex3D~\citep{han2025flex3d} is one of the few recent exceptions that explicitly performs candidate-view generation and curation, filtering views by quality and consistency before reconstruction. By contrast, active view selection or next-best-view (NBV) has been studied much more extensively in 3D \textit{reconstruction}.
Geometry-heuristic methods~\citep{pito1999nextbestview, border2018see, border2024see} choose views according to the current geometric state, for example, by prioritizing uncovered surfaces, missing regions, or occlusion boundaries. Information-gain methods~\citep{jiang2024fisherrf, wilson2025popgs} instead estimate how much uncertainty a candidate view is expected to reduce, often using probabilistic map uncertainty or Fisher-information-based criteria.
RL-based methods~\citep{peralta2020nbvpolicy, chen2024gennbv} learn view-selection policies in continuous spaces, and prediction-based methods~\citep{frahm2025vinnbv} directly estimate the reconstruction benefit of each candidate view. In contrast, we study a more challenging problem in real scenes: selecting anchor and helper views for severely occluded instances from highly redundant observations, so that helper views can provide missing information for 3D generation.


\section{More Details}\label{sec:implementation}
\textbf{Setup.} We evaluate our method on both the synthetic Replica dataset~\citep{straub2019replica} and the real-world ScanNet++ dataset~\citep{yeshwanth2023scannetpp}. 
Replica is a photorealistic indoor 3D scene dataset with dense mesh geometry and high-resolution textures. It contains 18 indoor scenes, including rooms and offices, and covers objects from 88 semantic categories.
It supports rendering multi-view RGB images from arbitrary camera poses. ScanNet++ is a real-world indoor dataset comprising 460 high-resolution 3D scenes and over 1,000 categories of 3D assets, together with DSLR images and RGB-D frames captured using an iPhone. 
Following previous work ~\citep{ni2025dprecon,kong2023vmap} that evaluates on both Replica and ScanNet++, we perform quantitative comparisons on selected subsets of both datasets.

\textbf{Implementation Details.} 
All experiments are conducted on a single NVIDIA H200 GPU with 140GB memory. 
We follow the default hyperparameter settings of SAM3D. 
For our proposed modules, we set the number of helper views to $M=3$ in Occupancy-Coverage View Selection, and the midpoint to $T_{\mathrm{mid}}=15$ in Step-Efficient Conflict-Aware Multiview Fusion. 
The balancing coefficients $\lambda_{d}$, $\lambda_{b}$, $\lambda_{p}$, and $\lambda_{r}$ in Rigid-Object Gaussian Layout Refinement are set to 2.0, 2.0, 0.04, and 0.05, respectively.
The balancing coefficients $\lambda_m$, $\lambda_{\mathrm{rgb}}$, $\lambda_d$, $\lambda_b$, $\lambda_c$, and $\lambda_p$ are set to $2.0$, $2.5$, $2.0$, $0.04$, $0.05$, and $0.5$, respectively.

\textbf{Metrics.} Following prior 3D object generation and scene reconstruction methods, we evaluate geometry quality at both object and scene levels using Chamfer Distance (CD), F-Score at a threshold of $0.05$, and Normal Consistency (NC). 
All instance-level Chamfer-based distances, including CD and CD$_{\mathrm{GT}\rightarrow\mathrm{Pred}}$, are reported in millimeters, while scene-level CD is reported in centimeters.
To further assess scene-level spatial arrangement, we report Mask IoU results. 
For appearance quality, we evaluate rendered reconstructions using PSNR, SSIM, and LPIPS at both object and scene levels. 
Since indoor scene annotations often provide incomplete instance meshes due to limited visibility and occlusion, symmetric CD may over-penalize plausible geometry completed in unobserved regions. 
We therefore additionally report the one-sided Chamfer Distance from ground truth to prediction, denoted as CD$_{\mathrm{GT}\rightarrow\mathrm{Pred}}$ or Completeness, which measures the coverage of observed ground-truth surfaces by the generated object mesh.


\textbf{Baselines.}
We compare Scene-SAM3D with both single-view and multi-view 3D generation baselines.
For single-view reconstruction, we include Trellis~\cite{trellis}, Amodal3R~\cite{wu2025amodal3r}, and SAM3D~\cite{sam3d}.
For multi-view reconstruction, we compare with InstantMesh~\cite{xu2024instantmesh}, ReconViaGen~\cite{reconviagen}, and a reproduced MultiDiffusion-based baseline, denoted as MD + SAM3D, which applies MultiDiffusion fusion~\cite{multidiffusion} on SAM3D~\cite{sam3d}.
For scene-level evaluation, we further compare different alignment strategies, including direct backprojection (BP) and ICP refinement~\cite{icp} initialized from BP, against our proposed Gaussian layout refinement.
All methods are evaluated under the same protocol, and multi-view methods use the same number of selected views for fair comparison.

\begin{figure}
    \centering
    \includegraphics[width=0.97\linewidth]{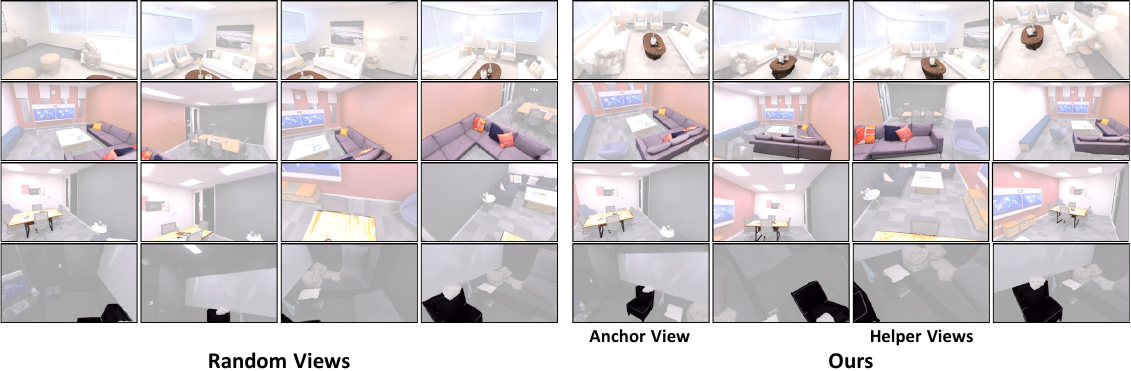}
    \caption{Selected views visualization on Replica.}
    \label{fig:sup_selected_views}
\end{figure}

\begin{figure}
    \centering
    \includegraphics[width=0.98\linewidth]{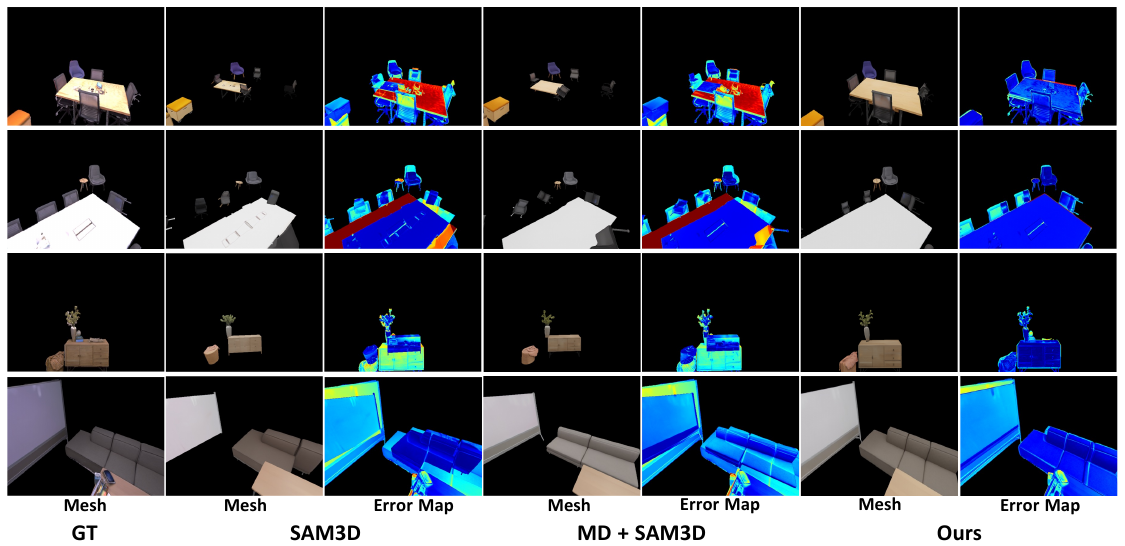}
    \caption{Qualitative visualization on Replica. In the error maps, warmer colors indicate larger alignment errors.}
    \label{fig:sup_replica}
\end{figure}

\begin{figure}
    \centering
    \includegraphics[width=0.98\linewidth]{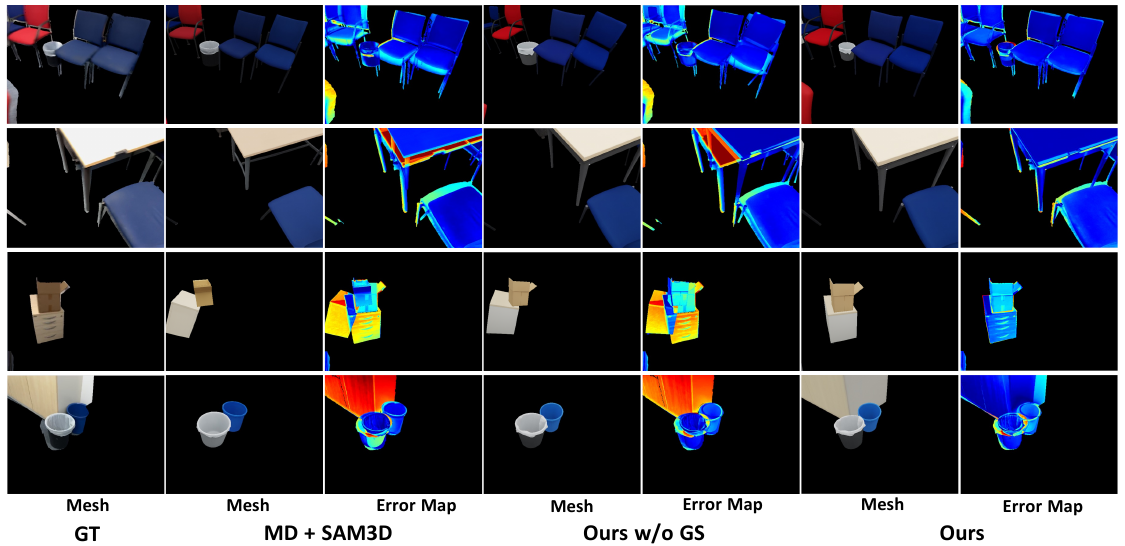}
    \caption{Qualitative visualization on ScanNet++. In the error maps, warmer colors indicate larger alignment errors.}
    \label{fig:sup_scannet}
\end{figure}

\section{More Visualization}
\label{sec:more_visualization}

We visualize the selected views by comparing our Occupancy-Coverage View Selection strategy with random view selection.
As shown in Fig.~\ref{fig:sup_selected_views}, random selection often produces incomplete or redundant observations, whereas our method selects a compact anchor-helper view set with more reliable and complementary 3D evidence.
Specifically, the anchor view maximizes world-space occupancy coverage, favoring views that explain a broader 3D extent of the target object, while the helper views further complement anchor-unobserved regions or provide additional local details.
For example, in the first-row table case, the randomly selected views observe the object only partially, whereas our selected views jointly cover the table more completely.
Similarly, in the third-row sofa case, random views lack sufficient observation of the sofa back, while our anchor and helper views complement each other to cover both the front and back sides.

We also provide additional qualitative results on Replica and ScanNet++ in Fig.~\ref{fig:sup_replica} and Fig.~\ref{fig:sup_scannet}, respectively, to better analyze geometry fidelity in the world coordinate system.
For each scene, we place the reconstructed objects back into the world coordinate system and render them from the same viewpoints as the ground truth.
We visualize both the rendered meshes and the corresponding surface error maps, where warmer colors indicate larger geometry errors.
The visualizations show that our method improves both object-level reconstruction quality and scene-level layout consistency.
First, our method produces more accurate object shapes and appearances, resulting in visibly lower surface errors on reconstructed objects.
For example, in Fig.~\ref{fig:sup_replica}, our method better preserves the geometry of the laundry basket in the third row and the sofa structure in the fourth row.
Similarly, in Fig.~\ref{fig:sup_scannet}, the bins in the fourth row are reconstructed with more faithful shape and appearance, leading to smaller error regions compared with the baselines.
Second, the error maps reveal that methods without our GS refinement often suffer from noticeable spatial offsets after placing objects back into the world coordinate system.
These errors include inaccurate object positions, rotations, and relative object-object arrangements.
For instance, in Fig.~\ref{fig:sup_replica}, the reconstructed table is placed too far from the surrounding objects, producing large errors in the scene layout.
In Fig.~\ref{fig:sup_scannet}, the cabinet and cardboard box in the third row exhibit an incorrect collision relationship before GS refinement.
After applying GS refinement, the reconstructed scene layout becomes much better aligned with the ground truth, as indicated by the reduced warm-colored regions in the error maps.
These results demonstrate that our method not only improves object geometry and appearance, but also produces more accurate world-space scene layouts.

\section{More Ablation Studies}
\label{sec: more_ablation}


\textbf{View Selection and Fusion Strategy.}
To evaluate our multi-view selection and fusion designs, we compare Scene-SAM3D with alternative view selection and fusion strategies.
For view selection, we compare our occupancy-coverage strategy with random selection and a k-means++ feature-diversity baseline~\cite{kmeans++}.
Specifically, the k-means++ baseline performs view selection in the multi-view feature space by favoring feature-diverse views, a criterion commonly used to reduce redundancy among candidate observations.
For fusion, we compare with stochastic view conditioning and MultiDiffusion fusion, both of which can be adapted to the 3D multi-view generation setting.
As shown in Tab.~\ref{tab:selectionfusion}, our method achieves the best overall performance in both view selection and fusion.
The only exception is SSIM, where different strategies show marginal differences.
This is because SSIM mainly measures image-space structural similarity and is less sensitive to localized 3D geometric artifacts, such as duplicated legs, discontinuous surfaces, or small missing parts.
Therefore, geometry-related metrics better reflect the advantages of our view selection and fusion strategy.

\begin{table*}[h]
    \centering
    \caption{Ablation on view selection and fusion strategy.}
    \label{tab:selectionfusion}
    \setlength{\tabcolsep}{6pt}
    \renewcommand{\arraystretch}{1.25}
    \newcolumntype{C}[1]{>{\centering\arraybackslash}p{#1}}
    \resizebox{1\textwidth}{!}{
    \begin{tabular}{l l C{1.0cm} C{1.0cm} C{1.0cm} C{1.3cm} C{1.0cm} C{1.0cm} C{1.0cm}}
        \toprule
     \multicolumn{2}{c}{} &
        \multicolumn{4}{c}{\textit{Geometry}} &
        \multicolumn{3}{c}{\textit{Reconstruction}} \\
        \cmidrule(lr){3-6} \cmidrule(lr){7-9}
        \textbf{Selection} &
        \textbf{Fusion} &
        CD$^\downarrow$ &
        Comp$^\downarrow$ &
        NC$^\uparrow$ &
        \makecell{F@0.05$^\uparrow$} &
        PSNR$^\uparrow$ &
        SSIM$^\uparrow$ &
        LPIPS$^\downarrow$ \\
        \midrule
        $\operatorname{random}$   & $\operatorname{Stochastic}$ & 46.35 & 38.34 & 74.10 & 0.68 & 13.06 & 0.71 & 0.43 \\
        $\operatorname{random}$   & $\operatorname{MultiDiffusion}$       & 45.83 & 37.66 & 74.24 & 0.69 & 13.09 & 0.71 & 0.42 \\
        $\operatorname{k\text{-}means\text{++}}$   & $\operatorname{Stochastic}$ & 42.45 & 33.56 & 74.76 & 0.71 & 13.07 & 0.71 & 0.41 \\
        $\operatorname{k\text{-}means\text{++}}$   & $\operatorname{MultiDiffusion}$       & 47.45 & 38.05 & 73.91 & 0.67 & 12.92 & 0.71 & 0.43 \\
        $\operatorname{coverage}$ & $\operatorname{MultiDiffusion}$       & 40.98 & 32.06 & 74.86 & 0.72 & 13.19 & 0.71 & 0.40 \\
        \rowcolor{headerblue}
        $\operatorname{coverage}$ & $\operatorname{Ours}$
            & \textbf{39.37} & \textbf{30.65} & \textbf{75.17} & \textbf{0.73}
            & \textbf{13.20} & 0.71
            & \textbf{0.40} \\
        \bottomrule
    \end{tabular}
    }
\end{table*}

\begin{table*}[ht]
    \centering
    \caption{Scene-level quantitative comparison on ScanNet++. As the 4-view Coverage strategy has been shown on Replica to consistently outperform both single-view inference and random view selection, we restrict the ScanNet++ comparison to the strongest baseline variants.}
    \label{tab:comparison_scannetpp}
    \renewcommand{\arraystretch}{1.25}
    \newcolumntype{C}[1]{>{\centering\arraybackslash}p{#1}}
    \resizebox{0.9\textwidth}{!}{
    \begin{tabular}{l c C{0.75cm} C{0.75cm} C{1.3cm} C{0.75cm} C{0.75cm} C{0.75cm} C{0.75cm}}
        \toprule
        \multicolumn{2}{c}{} &
        \multicolumn{3}{c}{\textit{Geometry}} &
        \multicolumn{4}{c}{\textit{Reconstruction}} \\
        \cmidrule(lr){3-5} \cmidrule(lr){6-9}
        \textbf{Method} &
        \textbf{Align} &
        CD$^\downarrow$ &
        NC$^\uparrow$ &
        \makecell{F@0.05$^\uparrow$} &
        \makecell{IoU$^\uparrow$} &
        PSNR$^\uparrow$ &
        SSIM$^\uparrow$ &
        LPIPS$^\downarrow$ \\
        \midrule
        $\operatorname{MD}$ \cite{multidiffusion}+$\operatorname{SAM3D}$ \cite{sam3d} & $\operatorname{BP}$  & 8.05 & 0.53 & 0.58 & 0.36 & 9.36 & 0.21 & 0.46 \\
        $\operatorname{MD}$ \cite{multidiffusion}+$\operatorname{SAM3D}$ \cite{sam3d} & $\operatorname{ICP}$ & 7.18 & 0.61 & 0.60 & 0.39 & 9.59 & 0.22 & 0.45 \\
        \midrule
        $\operatorname{Ours}$ & $\operatorname{ICP}$ & 6.89 & 0.63 & 0.61 & 0.40 & 9.65 & 0.22 & 0.44 \\
        \rowcolor{headerblue}
        $\operatorname{Ours}$ & $\operatorname{GS}$  & \textbf{4.76} & \textbf{0.74} & \textbf{0.69} & \textbf{0.57} & \textbf{11.07} & \textbf{0.31} & \textbf{0.37} \\
        \bottomrule
    \end{tabular}
    }
\end{table*}


\end{document}